\documentclass[lettersize, journal]{IEEEtran}
\usepackage{amsmath, amsfonts}
\usepackage[hidelinks]{hyperref}
\usepackage{algorithmic}
\usepackage{algorithm}
\usepackage{array}
\usepackage{subcaption}
\usepackage{textcomp}
\usepackage{stfloats}
\usepackage{url}
\usepackage{verbatim}
\usepackage{graphicx}
\usepackage{cite}
\usepackage{multirow}
\usepackage{booktabs}
\usepackage{amsmath}
\usepackage{wrapfig}
\usepackage{bbm}
\usepackage{mathtools}
\usepackage{authblk}
\usepackage{amsmath,amssymb,amsthm}
\usepackage[table, xcdraw]{xcolor}
\usepackage[T1]{fontenc}
\hypersetup{colorlinks, linkcolor={green!50!black}, citecolor={red!50!black}, urlcolor={blue!80!black}}
\usepackage[normalem]{ulem}

\begin{document}
    \newcommand{\gustavo}[1]{\textcolor{blue}{Gustavo: #1}}
    \newcommand{\fengbei}[1]{\textcolor{orange}{Fengbei: #1}}
    \newcommand{\chong}[1]{\textcolor{orange}{Chong: #1}}
    \newcommand{\RNum}[1]{\uppercase{\expandafter}{\romannumeral #1\relax}}

    \title{%
    
    Bridging Generative and Discriminative Noisy-Label Learning via Direction-Agnostic EM Formulation
    
    }
    \author{Fengbei Liu, Chong Wang, Yuanhong Chen, Yuyuan Liu, Gustavo Carneiro\thanks{Fengbei Liu is with Cornell Tech, Cornell University. Chong Wang is with Department of Radiology, Stanford University. Yuanhong Chen is with Australian Institute for Machine Learning (AIML), University of Adelaide. Yuyuan Liu is with Oxford University. Gustavo Carneiro is with Center of Vision, Speech and Signal Processing (CVSSP), University of Surrey. \textit{(Fengbei Liu and Chong Wang contributed equally to this work)}. \textit{(Corresponding author: Gustavo Carneiro)}. Manuscript submitted Dec 2023; Manuscript revised Oct 2025; Manuscript revised Dec 2025.}}

    \markboth{IEEE TRANSACTIONS ON PATTERN ANALYSIS AND MACHINE INTELLIGENCE,~Vol.~14, No.~8, March~2026}%
    {Shell \MakeLowercase{\textit{et al.}}: A Sample Article Using IEEEtran.cls for IEEE Journals}

    \IEEEpubid{0000--0000/00\$00.00~\copyright~2021 IEEE}

    \maketitle
    \begin{abstract}
        
        Although noisy-label learning is often approached with discriminative methods for simplicity and speed, generative modeling offers a principled alternative by capturing the joint mechanism that produces features, clean labels, and corrupted observations. However, prior work typically (i) introduces extra latent variables and heavy image generators that bias training toward reconstruction, (ii) fixes a single data-generating direction (\(Y\rightarrow\!X\) or \(X\rightarrow\!Y\)), limiting adaptability, and (iii) assumes a uniform prior over clean labels, ignoring instance-level uncertainty. Here, we propose a single-stage, EM-style framework for generative noisy-label learning that is \emph{direction-agnostic} and avoids explicit image synthesis. First, we derive a single Expectation-Maximization (EM) objective whose E-step specializes to either causal orientation without changing the overall optimization objective. Second, we replace the intractable \(p(X\mid Y)\) with a dataset-normalized discriminative proxy computed using a discriminative classifier on the finite training set, retaining the structural benefits of generative modeling at much lower cost. Third, we introduce \emph{Partial-Label Supervision} (PLS), an instance-specific prior over clean labels that balances coverage and uncertainty, improving data-dependent regularization. Across standard vision and natural language processing (NLP) noisy-label benchmarks, our method achieves state-of-the-art accuracy, lower transition-matrix estimation error, and substantially less training computation than current generative and discriminative baselines. Code: \url{https://github.com/lfb-1/EMNL}.

    \end{abstract}

    \section{Introduction}
    Deep neural networks (DNNs) have achieved remarkable success in computer vision~\cite{he2015deep,krizhevsky2017imagenet}, natural language processing (NLP)~\cite{devlin2018bert,young2018recent}, and medical image analysis~\cite{litjens2017survey,wang2017chestx}. However, DNNs often require large amounts of high-quality annotated data for supervised training~\cite{deng2009imagenet}, which is challenging and expensive to acquire. To alleviate this problem, some datasets have been annotated via crowdsourcing~\cite{xiao2015learning}, search engines~\cite{song2019selfie}, or via NLP applied to radiology reports~\cite{wang2017chestx}. While these lower-cost annotation processes enable the construction of large-scale datasets, they inevitably introduce noisy labels, leading to performance degradation. Therefore, novel learning algorithms are needed to robustly train DNNs when training sets contain noisy labels, forming a noisy-label learning task. The main challenge in noisy-label learning is that we observe only the data, represented by a random variable $X$, and the corresponding noisy label, denoted by the variable $\tilde{Y}$, yet we aim to estimate the model $p(Y \mid X)$, where $Y$ is the unobserved clean-label variable.

    \IEEEpubidadjcol
    Most existing methods proposed in the field resort to two types of discriminative learning strategies, namely sample selection and noise-transition modeling. \textit{Sample selection}~\cite{arazo2019unsupervised,han2018co,li2020dividemix} trains the model $p_{\theta}(Y \mid X)$, parameterized by $\theta$, by maximizing the likelihood over pseudo-clean training samples, as follows:

    \begin{equation}
        \begin{aligned}
            \theta^{*}= \arg \max_{\theta}\mathbb{E}_{P(X,\tilde{Y})}\left [ \mathsf{clean}(X,\tilde{Y}) \times p_{\theta}(\tilde{Y}| X) \right ],                        \\
            \text{ where }\mathsf{clean}(X=\mathbf{x},\tilde{Y}=\tilde{\mathbf{y}}) = \begin{cases}1 \text{, if } Y=\tilde{\mathbf{y}} \\ 0 \text{, otherwise}\end{cases}
        \end{aligned}
        \label{eq:optimisation_discriminative_sample_selection}
    \end{equation}
    and $P(X,\tilde{Y})$ denotes the distribution used to generate the training set’s data points and noisy labels. Note that $\mathbb{E}_{P(X,\tilde{Y})}\left [ \mathsf{clean}(X,\tilde{Y}) \times p_{\theta}(\tilde{Y}| X) \right ] \equiv \mathbb{E}_{P(X,Y)}\left [ p_{\theta}(Y | X) \right ]$ if the function $\mathsf{clean}(\cdot)$ successfully selects the clean-label training samples. Unfortunately, $\mathsf{clean}(\cdot)$ usually relies on the \textit{small-loss hypothesis}~\cite{arpit2017closer} to select the $R \%$ smallest-loss training samples, which offers little guarantee of successfully selecting clean-label samples. Approaches based on the noise \textit{transition matrix}~\cite{xia2020part,cheng2022instance,patrini2017making} aim to estimate a clean-label classifier and the label transition, as follows:
    \begin{equation}
        \begin{split}
            \theta^{*}&=\arg\max_{\theta}\mathbb{E}_{P(X,\tilde{Y})}\left [ \sum_{Y}p(\tilde{Y},Y|X) \right ] \\
            &=\arg\max_{\theta_1,\theta_2}\mathbb{E}_{P(X,\tilde{Y})}\left [ \sum_{Y}p_{\theta_1}(\tilde{Y}|Y,X) p_{\theta_2}(Y|X) \right ],
        \end{split}
        \label{eq:optimisation_discriminative_transition}
    \end{equation}
    where $\theta = [\theta_{1},\theta_{2}]$, $p_{\theta_1}(\tilde{Y}|Y,X)$ represents a label-transition matrix, often simplified to be class-dependent with $p_{\theta_1}(\tilde{Y}| Y) = p_{\theta_1}(\tilde{Y}| Y,X)$. Since we do not have access to the label-transition matrix, we need to estimate it from the noisy-label training set, which is challenging because of identifiability issues~\cite{liu2022identifiability}, making the use of anchor points~\cite{patrini2017making} or other types of regularization~\cite{cheng2022instance} necessary.

    \begin{figure}
        \centering
        \includegraphics[width=0.9\linewidth]{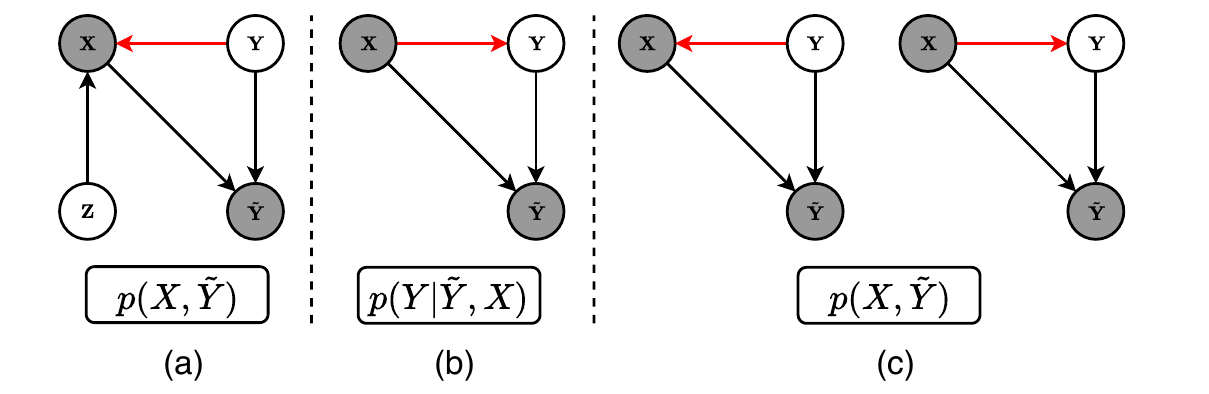}
        \vspace{-5pt}
        \caption{
        Generative noisy-label learning models and their corresponding probability functions, where the red arrow indicates the causal direction between $X$ and $Y$. \textbf{(a)} CausalNL~\cite{yao2021instance} and InstanceGM~\cite{garg2023instance} assume that $Y \rightarrow X$ and optimize the joint likelihood $p(X,\tilde{Y})$, which requires an additional latent variable $Z$ for generation. \textbf{(b)} NPC~\cite{bae2022noisy} and DyGEN~\cite{zhuang2023dygen} assume that $X \rightarrow Y$ and optimize $p(Y|\tilde{Y},X)$ as a post-processing step; they do not require $Z$. \textbf{(c)} Our proposed method optimizes $p(X,\tilde{Y})$ and accommodates both causal directions across datasets, without the need to model $Z$.
        }
        \label{fig:causal_relationships}
    \end{figure}

    On the other hand, generative noisy-label (GNL) learning~\cite{bae2022noisy,yao2021instance,zhuang2023dygen} aims to estimate the transition matrix by assuming a causal relationship between $X$ and $Y$, as depicted in Fig.~\ref{fig:causal_relationships}. Previous methods consider a causal generative process where $Y\rightarrow X$~\cite{yao2021instance,garg2023instance} and are trained to maximize the joint data likelihood $p(\tilde{Y},X) = \int_{Y,Z}p(X|Y,Z)p(\tilde{Y}|Y,X)p(Y)p(Z)dYdZ$, where $Z$ denotes a low-dimensional latent image representation and $Y$ is the latent clean label. This optimization requires a variational distribution $q_{\phi}(Y,Z|X)$ to maximize the evidence lower bound (ELBO):
    \begin{equation}
        \scalebox{0.75}{$\begin{split}
            \theta_{1}^{*},\theta_{2}^{*},\phi^{*}=\arg\max_{\theta_1,\theta_2,\phi}\mathbb{E}_{q_{\phi}(Y,Z|X)}\left[ \log \left ( \frac{p_{\theta_1}(X|Y,Z)p_{\theta_2}(\tilde{Y}|Y,X)p(Y)p(Z)}{q_{\phi}(Y,Z|X)}\right ) \right ],
        \end{split}$}\label{eq:optimisation_generative}
    \end{equation}
    where $p_{\theta_1}(X|Y,Z)$ denotes an image generative model, $p_{\theta_2}(\tilde{Y}|Y,X)$ represents the label-transition model, $p(Z)$ is the latent image representation prior (commonly assumed to be a standard normal distribution), and $p(Y)$ is the clean-label prior, which is usually assumed to be a uniform distribution. Note from Eq.~\eqref{eq:optimisation_generative} that the optimization of the transition matrix $p_{\theta_2}(\tilde{Y}|Y,X)$ is constrained by the training of $p_{\theta_1}(X|Y,Z)$. This constrained optimization can reduce uncertainty in the transition matrix, suggesting that optimizing this generative model naturally improves transition-matrix estimation without requiring the regularization strategies explored by discriminative models~\cite{li2021provably,xia2019anchor,cheng2022instance,patrini2017making}.

    However, GNL faces several limitations that hinder its broader adoption. For instance, current methods rigidly assume a particular causal direction between $X$ and $Y$~\cite{bae2022noisy,zhuang2023dygen,yao2021instance,garg2023instance} and cannot easily adapt to alternative causal directions~\cite{pmlr-v202-yao23a}. Moreover, introducing a latent variable $Z$ requires an image-generation module (e.g., a variational autoencoder~\cite{kingma2013auto}), which limits the framework to low-resolution images, increases complexity, and results in suboptimal, reconstruction-driven training that degrades performance. Although some generative methods do not use $Z$ and instead rely on a post-processing framework to calibrate pretrained classifier predictions~\cite{bae2022noisy,zhuang2023dygen}, they assume that high-confidence classification predictions from $p(\tilde{Y}\mid X)$ are reliable indicators of the true labels $Y$, a heuristic that may not hold for highly noisy datasets. Thus, GNL still requires a principled approach to unify these different optimization goals.

    To address the limitations of existing GNL approaches, we propose a novel, single-stage GNL framework—similar to discriminative approaches—that is adaptable to different causal processes and is based on the Expectation–Maximization (EM) framework. First, we derive the optimization objective under the EM formulation and show that different causal directions, $Y \rightarrow X$ or $X \rightarrow Y$, can be handled under the same objective without substantially changing the architecture. Second, we remove the latent variable $Z$ and approximate the typically generative term $p(X\mid Y)$ using the output of a discriminative classifier. Finally, because removing $Z$ implies that $X$ should be constrained by an informative $Y$, we propose a new partial-label supervision (PLS) training signal for noisy-label learning that considers both clean-label coverage and label uncertainty. This instance-wise prior regularizes training by reducing $p(Y)$ to a one-hot distribution when the model has high confidence that the training label is correct (i.e., low label uncertainty); conversely, $p(Y)$ is approximated by a uniform distribution when the model has low confidence that the training label is clean (i.e., high label uncertainty). We summarize our contributions as follows:
    
    \begin{itemize}
        
        \item \textbf{Unified EM-based formulation}: We reformulate the optimization objective of generative noisy-label learning under the EM framework, enabling different causal directions to be implemented within a unified optimization process.
        \item \textbf{Efficient approximation of $p(X\mid Y)$}: We approximate $p(X\mid Y)$ using a finite set of training samples and the output of a discriminative classifier, avoiding explicit estimation of the latent variable $Z$ while preserving the benefits of generative modeling.
        \item \textbf{Partial-Label Supervision (PLS)}: We introduce a novel PLS strategy for noisy-label learning that dynamically approximates $Y$ using an informative partial-label prior $p(Y)$, accounting for both latent clean-label coverage and uncertainty.
        
    \end{itemize}
    We conduct extensive experiments on both synthetic and real-world noisy-label benchmarks for computer vision and natural language processing (NLP) tasks. Experimental results and comprehensive ablation studies show that our framework outperforms previous generative approaches~\cite{bae2022noisy,zhuang2023dygen,yao2021instance,garg2023instance} in terms of performance, computational cost, and transition-matrix estimation error. Comparisons with label-transition matrix approaches~\cite{dualT2020nips,bae2022noisy} show that our approach yields more accurate classification and transition-matrix estimation without imposing additional regularization to handle the identifiability problem~\cite{yao2021instance}. Regarding sample-selection methods~\cite{li2020dividemix,garg2023instance}, our approach achieves competitive results at significantly lower computational costs.

    \section{Related Work}
    \label{sec:related_work}

    \subsection{Discriminative noisy-label learning}

    \textbf{Sample selection} is a discriminative noisy-label learning strategy, defined in Eq.~\eqref{eq:optimisation_discriminative_sample_selection}, which must address two problems: 1) the definition of the function $\mathsf{clean}(\cdot)$, and 2) what to do with the samples classified as noisy. Most definitions of $\mathsf{clean}(\cdot)$ classify small-loss samples~\cite{arpit2017closer} as pseudo-clean~\cite{arazo2019unsupervised,chen2019understanding,han2018co,jiang2017mentornet,li2020dividemix,malach2017decoupling,shen2019learning,wei2020combating,li2023disc}. Other approaches select clean samples based on $K$-nearest neighbor classification from deep learning feature spaces~\cite{ortego2021multi,wang2018iterative}, distance to the class-specific eigenvector from the Gram-matrix eigendecomposition using intermediate deep learning feature spaces~\cite{kim2021fine}, uncertainty measures~\cite{kohler2019uncertainty}, or prediction consistency between teacher and student models~\cite{kaiser2022blind}. After sample classification, some methods discard the noisy-label samples during training~\cite{chen2019understanding,jiang2017mentornet,malach2017decoupling,shen2019learning}, while others use them for semi-supervised learning~\cite{li2020dividemix}.
    The main issue with these strategies is the hard-decision rule that classifies samples as having clean or noisy labels, as well as pseudo-labeling noisy samples with single-model predictions, leaving little room for uncertainty about possible label candidates. %
    These strategies often involve high training complexity, stemming from the use of semi-supervised learning and multiple models.

    \textbf{Label-transition model} is another discriminative noisy-label learning strategy in Eq.~\eqref{eq:optimisation_discriminative_transition} that depends on a reliable estimate of $p(\tilde{Y}|Y,X)$~\cite{cheng2022instance,patrini2017making,xia2020part}. Forward-T~\cite{patrini2017making} uses an additional classifier and anchor points from clean-label samples to learn a class-dependent transition matrix. Part-T~\cite{xia2020part} estimates an instance-dependent model. MEDITM~\cite{cheng2022instance} uses manifold regularization to estimate the label-transition matrix. In general, estimating this label-transition matrix is under-constrained, leading to the identifiability problem~\cite{liu2022identifiability}, which is addressed with strong assumptions~\cite{patrini2017making} or with additional labels from each training sample~\cite{liu2022identifiability}.

    \subsection{Generative modeling in noisy-label learning}
    Generative modeling is a noisy-label learning technique~\cite{bae2022noisy,garg2023instance,yao2021instance,zhuang2023dygen} that explores different graphical models (see Fig.~\ref{fig:causal_relationships}) to estimate the clean label of an image.
    Specifically, CausalNL~\cite{yao2021instance} and InstanceGM~\cite{garg2023instance} assume that the latent clean label $Y$ generates $X$, and the noisy-label $\tilde{Y}$ is generated from $X$ and $Y$. The drawback of CausalNL and InstanceGM is that they optimize the joint distribution of data and noisy-labels $p(X, \tilde{Y})$ instead of directly optimizing a model that associates the data with the clean label, such as $p(X | Y)$ or $p(Y | X)$. They also introduce the random variable $Z$ for image generation, which is computationally expensive. NPC~\cite{bae2022noisy} and DyGEN~\cite{zhuang2023dygen} both assume that $X$ generates $Y$ and propose a post-processing technique for inferring the clean label $Y$. However, they rely on the premise that high-confidence predictions by the noisy classifier can be used as a prior for $Y$, which may not be effective in datasets with a high rate of label noise. Moreover, the above methods are designed for a fixed causal direction, 
    and may not adapt well to datasets with different causal relationships.
    
    \subsection{Weak supervision in noisy-label learning }
    
    Because clean labels $Y$ are unavailable, noisy-label learning often involves estimating a \textbf{clean-label prior distribution} $p(Y)$ to provide the model with a reliable supervisory signal.
    Conventional techniques include Mixup~\cite{zhang2017mixup}, label smoothing~\cite{Lukasik_ICML_2020_label_smoothing_label_noisy}, pseudo-labeling~\cite{qiu2021source,lin2022prototype,li2024nicest,liu2022perturbed}, and relabeling~\cite{li2020dividemix,liu2022acpl}, which are commonly used in noisy-label learning, unsupervised domain adaptation, and scene graph generation. Ideally, the design principles for producing $p(Y)$ are: 1) increasing clean-label coverage, and 2) reducing the uncertainty of the label prior. However, Mixup and label smoothing are effective approaches for proposing soft labels for noisy-label learning, increasing label coverage but not necessarily reducing label uncertainty. Pseudo-labeling and relabeling replace the training label with a hard pseudo-label from model predictions, making them effective at reducing uncertainty but offering limited improvement in coverage.

    \textbf{Partial-label learning (PLL)}  offers a clear trade-off between coverage and uncertainty. It assumes that each image is associated with a candidate label set, called a partial label~\cite{tian2023partial}, where one of the labels is a true positive and the remaining labels are false positives. The goal of PLL is to predict the single true label associated with each training sample. %
    PICO~\cite{wang2022pico} uses contrastive learning within an EM optimization to address PLL. CAV~\cite{zhang2021exploiting} proposes class activation mapping to identify the true label within the candidate set. PRODEN~\cite{lv2020progressive} progressively identifies the true labels from a candidate set to update the model parameters. Qiao et al.~\cite{qiao2022decompositional} explicitly model the generation process of candidate labels with correct and incorrect label generation. Recent work called NPN~\cite{sheng2024adaptive} decomposes the label space into candidate labels and non-candidate labels, and employs partial-label learning on the candidate label set and negative learning on the remaining non-candidate label set. However, to avoid trivial solutions it relies on negative learning instead of modeling the uncertainty of partial labels. Furthermore, it requires consistency regularization and heavy data augmentation to achieve satisfactory performances. The design of our PLS, which produces a prior label distribution represented by $p(Y)$, is inspired by PLL, but unlike PLL, our $p(Y)$ is dynamically constructed during training since the clean label is latent. In particular, $p(Y)$ will approximate a uniform distribution if the training label has a high probability of being wrong. On the other hand, $p(Y)$ will condense to a one-hot distribution when the training label has a high probability of being correct.

    \section{Method}
    \label{sec:methods}

    We denote the noisy training set as $\mathcal{D}= \{ (\mathbf{x}_{i},\tilde{\mathbf{y}}_{i})\}_{i=1}^{|\mathcal{D}|}$, where $\mathbf{x}_{i}\in \mathcal{X}\subset \mathbb{R}^{H \times W \times C}$ is the input image of size $H \times W$ with $C$ color channels, and $\tilde{\mathbf{y}}_{i}\in \mathcal{Y}\subset \{0,1\}^{|\mathcal{Y}|}$ is the observed one-hot noisy-label of the $i^{th}$ sample. We also denote $\mathbf{y}_{i} \in \mathcal{Y}$ as the latent clean label of the $i^{th}$ sample.
    Below, in Sec.~\ref{sec:framework} and~\ref{sec:pyx_optim}, we introduce the optimization goal of our model under different causal relationships. In Sec.~\ref{sec:generative_app}, we discuss how to approximate the generative term without the latent variable denoted by $Z$. In Sec.~\ref{sec:construct_partial}, we describe how to construct the informative prior using the partial-label strategy. The overall training algorithm is detailed in Sec.~\ref{sec:training}.

    \subsection{Formulation of the joint likelihood}
    We first consider the joint likelihood of $\mathbf{x}$ and the noisy-label $\tilde{\mathbf{y}}$, and our objective is to maximize $\log p(\mathbf{x}, \tilde{\mathbf{y}})$, which can be decomposed as:
    \begin{equation}
    \log p(\mathbf{x}, \tilde{\mathbf{y}})= \log \frac{p(\mathbf{x}, \mathbf{y}, \tilde{\mathbf{y}})}{p(\mathbf{y} | \mathbf{x}, \tilde{\mathbf{y}})}.
    \label{eq:optim_joint_p_x_tildey}
    \end{equation}
    The optimization of the log-likelihood in Eq.~\eqref{eq:optim_joint_p_x_tildey} can be achieved by introducing a variational posterior $q(\mathbf{y}| \mathbf{x})$, with:

    \begin{equation}
    \scalebox{0.9}{$
        \begin{split}
            \log p(\mathbf{x}, \tilde{\mathbf{y}})&= \mathbb{E}_{q(\mathbf{y} | \mathbf{x})}\left [ \log\frac{p(\tilde{\mathbf{y}}, \mathbf{y} , \mathbf{x})}{q(\mathbf{y} | \mathbf{x})}\right ] + \mathbb{E}_{q(\mathbf{y} | \mathbf{x})}\left [ \log\frac{q(\mathbf{y} | \mathbf{x})}{p(\mathbf{y} | \mathbf{x}, \tilde{\mathbf{y}})}\right ], \\
        \end{split}
        $}
        \label{eq:optim_joint_p_x_tildey2}
    \end{equation}
    where the second term can be expanded as follows:
    \begin{equation}
    \scalebox{0.9}{$
        \begin{split}
            \mathbb{E}_{q(\mathbf{y} | \mathbf{x})}\left [ \log\frac{q(\mathbf{y} | \mathbf{x})}{p(\mathbf{y} | \mathbf{x}, \tilde{\mathbf{y}})}\right ]&= \mathbb{E}_{q(\mathbf{y} | \mathbf{x})}\left [ \log\frac{q(\mathbf{y} | \mathbf{x})p(\mathbf{x}, \tilde{\mathbf{y}})}{p(\mathbf{x}, \mathbf{y}, \tilde{\mathbf{y}})}\right ]\\
            &=\mathbb{E}_{q(\mathbf{y} | \mathbf{x})}\left [ \log\frac{q(\mathbf{y} | \mathbf{x})p(\mathbf{x}, \tilde{\mathbf{y}})}{p(\tilde{\mathbf{y}} | \mathbf{x}, \mathbf{y})p(\mathbf{x} | \mathbf{y}) p(\mathbf{y})}\right ].
        \end{split}
        $}
        \label{eq:KL_Y_to_X}
    \end{equation}
    The formulation in Eq.~\eqref{eq:KL_Y_to_X} allows us to 
    derive two alternative causal formulations: the causal generation process $Y \rightarrow X$, denoted by $p(\mathbf{x}| \mathbf{y})$, and the anti-causal generation process $X \rightarrow Y$, represented by $p(\mathbf{y}| \mathbf{x})$. We now detail the formulation of each causal generation process below.

    \subsection{Formulation for the $Y \rightarrow X$ relationship}
    \label{sec:framework} 
    
    Substituting Eq.~\eqref{eq:KL_Y_to_X} into Eq.~\eqref{eq:optim_joint_p_x_tildey2} yields: 
    \begin{equation}
        \begin{split}
            \log p(&\mathbf{x},\tilde{\mathbf{y}}) = \\
            &\mathbb{E}_{q(\mathbf{y} | \mathbf{x})}\left [ \log\frac{p(\tilde{\mathbf{y}}, \mathbf{y} , \mathbf{x})}{q(\mathbf{y} | \mathbf{x})}\right ] + \mathbb{E}_{q(\mathbf{y} | \mathbf{x})}\left [ \log\frac{q(\mathbf{y} | \mathbf{x})}{p(\tilde{\mathbf{y}} | \mathbf{y}, \mathbf{x})p(\mathbf{y})}\right ]\\
            &-\mathbb{E}_{q(\mathbf{y} | \mathbf{x})}[\log p(\mathbf{x}| \mathbf{y})] + \log p(\mathbf{x}, \tilde{\mathbf{y}}).
        \end{split}
    \end{equation}
    Rearranging the terms, we obtain
    \begin{equation}
        \begin{split}                       &\mathbb{E}_{q(\mathbf{y} | \mathbf{x})}[\log p(\mathbf{x}| \mathbf{y})] = \\  &\;\;\underbrace{\mathbb{E}_{q(\mathbf{y} | \mathbf{x})}\left [ \log\frac{p(\tilde{\mathbf{y}}, \mathbf{y} , \mathbf{x})}{q(\mathbf{y} | \mathbf{x})}\right ]}_{\mathcal{L}(q,p)} + 
        \underbrace{\mathbb{E}_{q(\mathbf{y} | \mathbf{x})}\left [ \log\frac{q(\mathbf{y} | \mathbf{x})}{p(\tilde{\mathbf{y}} | \mathbf{y}, \mathbf{x})p(\mathbf{y})}\right ]}_{\mathsf{KL}[q\|p]}, 
        \end{split}
        \label{eq:pxy_EM}
    \end{equation}
    or equivalently,
\begin{equation}
    \mathcal{L}(q,p) = \mathbb{E}_{q(\mathbf{y} | \mathbf{x})}[\log p(\mathbf{x}| \mathbf{y})] - \mathsf{KL}[q\|p].
\end{equation}    
    This expression reveals a fundamental trade-off: maximizing the data-likelihood term $\mathbb{E}_{q(\mathbf{y} | \mathbf{x})}[\log p(\mathbf{x}| \mathbf{y})]$ inherently depends on minimizing the divergence between the approximate posterior $q(\mathbf{y}|\mathbf{x})$ and the true model distribution $p(\mathbf{y},\tilde{\mathbf{y}}|\mathbf{x})$. In practice, instead of optimizing $\mathbb{E}_{q(\mathbf{y} | \mathbf{x})}[\log p(\mathbf{x}| \mathbf{y})]$ directly, we maximize its evidence lower bound (ELBO):
    \begin{equation}
        \mathcal{L}(q,p) \le \mathbb{E}_{q(\mathbf{y} | \mathbf{x})}[\log p(\mathbf{x}| \mathbf{y})].
        \label{eq:ELBO}
    \end{equation}
    Because the KL divergence is nonnegative, the bound is tight when we minimize $\mathsf{KL}[q\|p]$.    
    
    Moreover, the ELBO $\mathcal{L}(q,p)$ in Eq.~\eqref{eq:pxy_EM} can be further
    decomposed into:
    \begin{equation}
        \begin{split}
            \mathbb{E}_{q(\mathbf{y} | \mathbf{x})}\left[ \log\frac{p(\tilde{\mathbf{y}}, \mathbf{y}, \mathbf{x})}{q(\mathbf{y} | \mathbf{x})}\right] =&%
            \mathbb{E}_{q(\mathbf{y}| \mathbf{x})}\left[ \log p(\tilde{\mathbf{y}}| \mathbf{x}, \mathbf{y}) \right] \\
            &+ \mathbb{E}_{q(\mathbf{y}| \mathbf{x})}\left[ \log \frac{p(\mathbf{x}|\mathbf{y})p(\mathbf{y})}{q(\mathbf{y} | \mathbf{x})}\right]. \\
        \end{split}
        \label{eq:two_expectation}
    \end{equation}
    Based on Eq.~\eqref{eq:pxy_EM} and Eq.~\eqref{eq:two_expectation}, the expected log of $p(\mathbf{x}|\mathbf{y})$ is defined as:
    \begin{equation}
        \begin{split}
            \mathbb{E}_{q(\mathbf{y} | \mathbf{x})}&\left [\log p(\mathbf{x}| \mathbf{y}) \right ] = \\
            &\underbrace{\mathbb{E}_{q(\mathbf{y}| \mathbf{x})} \left[ \log p(\tilde{\mathbf{y}}| \mathbf{x} , \mathbf{y}) \right] - \mathsf{KL}\left[q(\mathbf{y} | \mathbf{x}) \| p(\mathbf{x}|\mathbf{y})p(\mathbf{y}) \right]}_{\text{M-step}}\\
            &\underbrace{+ \mathsf{KL}\left[q(\mathbf{y} | \mathbf{x}) \|p(\tilde{\mathbf{y}} | \mathbf{x}, \mathbf{y})p(\mathbf{y}) \right]}_{\text{E-step}},
        \end{split}
        \label{eq:optim_goal}
    \end{equation}
    where the optimization is performed by minimizing the KL divergence in the E-step and maximizing the ELBO in the M-step.

    \subsection{Formulation for the $X \rightarrow Y$ relationship}
    \label{sec:pyx_optim} A similar principle applies to the anti-causal generation process $X \rightarrow Y$, where we apply a slightly different expansion as the one in Eq.~\eqref{eq:KL_Y_to_X}, with:
    \begin{equation}
    \scalebox{0.9}{$
        \begin{split}
            \mathbb{E}_{q(\mathbf{y} | \mathbf{x})}\left [ \log\frac{q(\mathbf{y} | \mathbf{x})p(\mathbf{x}, \tilde{\mathbf{y}})}{p(\mathbf{x}, \mathbf{y}, \tilde{\mathbf{y}})}\right ]&= \mathbb{E}_{q(\mathbf{y} | \mathbf{x})}\left [ \log\frac{q(\mathbf{y} | \mathbf{x})p(\mathbf{x}, \tilde{\mathbf{y}})}{p(\tilde{\mathbf{y}} | \mathbf{x}, \mathbf{y})p(\mathbf{y} | \mathbf{x}) p(\mathbf{x})}\right ].            
        \end{split}
        $}
    \end{equation}
    This leads to a similar formulation as the one in Eq.~\eqref{eq:optim_goal}, with a slight difference in the E-step:

    \begin{equation}
        \begin{split}
            \mathbb{E}_{q(\mathbf{y} | \mathbf{x})}&\left [\log p(\mathbf{y}| \mathbf{x}) \right ] = \\
            &\underbrace{\mathbb{E}_{q(\mathbf{y}| \mathbf{x})} \left[ \log p(\tilde{\mathbf{y}}| \mathbf{x} , \mathbf{y}) \right] - \mathsf{KL}\left[q(\mathbf{y} | \mathbf{x}) \| p(\mathbf{x}|\mathbf{y})p(\mathbf{y}) \right]}_{\text{M-step}}\\
            &\underbrace{+ \mathsf{KL}\left[q(\mathbf{y} | \mathbf{x}) \|p(\tilde{\mathbf{y}} | \mathbf{x}, \mathbf{y})
            p(\mathbf{x} )
            \right]}_{\text{E-step}},
        \end{split}
        \label{eq:pyx_optim_goal}
    \end{equation}
    where training is again based on minimizing the KL divergence in the E-step and maximizing the ELBO in the M-step.

    It is important to note that the above formulation gives our framework the flexibility to align the training procedure with the proposed causal or anti-causal structures while preserving EM's convergence guarantee.
    Although we have derived the overall optimization objective, some components remain difficult to estimate, in particular:
    \begin{itemize}
        \item $p(\mathbf{x}| \mathbf{y})$ and $p(\mathbf{x})$ in Eq.~\eqref{eq:optim_goal} and Eq.~\eqref{eq:pyx_optim_goal}: these terms are generative, and prior work typically estimates them with dedicated generative modules.

        \item $p(\mathbf{y})$ in Eq.~\eqref{eq:optim_goal}: this term is the clean-label prior and remains challenging in noisy-label learning because only noisy labels are present in the training set. Prior work typically approximates it using heuristics.
    \end{itemize}
We propose solutions for estimating these terms in the sections that follow.

    \begin{figure*}
        \centering
        \includegraphics[width=1.0\linewidth]{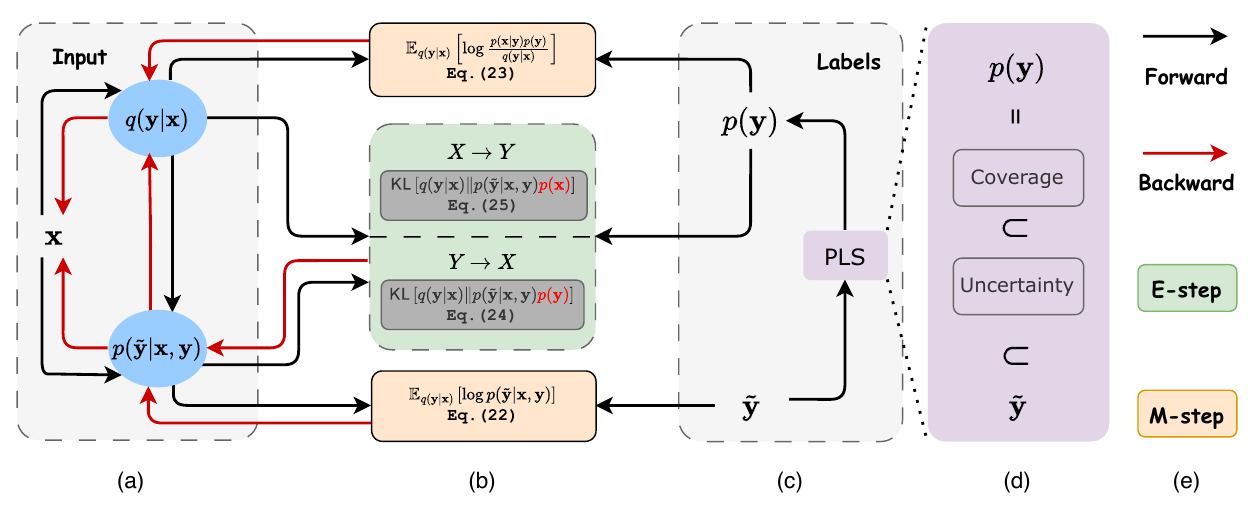}
        \vspace{-20pt}
        \caption{Description of the proposed framework. \textbf{(a)} shows the input image $\mathbf{x}$ and the model components. The term $q(\mathbf{y}|\mathbf{x})$ is the variational posterior in Eq.~\eqref{eq:optim_joint_p_x_tildey2} from the clean-label classifier, and $p(\tilde{\mathbf{y}}| \mathbf{y}, \mathbf{x})$ is the noise-transition module in Eq.~\eqref{eq:two_expectation}. \textbf{(b)} presents our Expectation--Maximization (EM) objective introduced in Eq.~\eqref{eq:optim_goal} and Eq.~\eqref{eq:pyx_optim_goal} and the corresponding implementation losses. In the E-step (green), we minimize one of the $KL$ divergences (Eq.~\eqref{eq:KL_divergence_pxy} or Eq.~\eqref{eq:KL_divergence_pyx}), depending on whether the data-generating process is causal or anti-causal. In the M-step (orange), we maximize the ELBO and the associated expectations by minimizing the losses in Eq.~\eqref{eq:ce} and Eq.~\eqref{eq:partial_loss}, respectively. \textbf{(c)} illustrates the labels used for supervision. $\tilde{\mathbf{y}}$ denotes the original noisy label. $p(\mathbf{y})$ is the constructed clean-label prior. ``PLS'' denotes the proposed Partial-Label Supervision introduced in Sec.~\ref{sec:construct_partial}. \textbf{(d)} shows the pipeline of constructing $p(\mathbf{y})$, where ``Coverage'' and ``Uncertainty'' correspond to the criteria defined in Eq.~\eqref{eq:moving_average} and Eq.~\eqref{eq:min_ineff}, respectively. \textbf{(e)} illustrates the legend of the figure, where black arrows represent the forward pass while red arrows denote the backward pass.}  %
        \label{fig:architectures}
    \end{figure*}
    \subsection{Approximation of $p(\mathbf{x}|\mathbf{y})$ and $p(\mathbf{x})$}
    \label{sec:generative_app}
    The terms $p(\mathbf{x}| \mathbf{y})$ and $p(\mathbf{x})$ are intractable to compute or normalize exactly. 
    One solution to mitigate this challenge is to use a latent image representation $Z$ to ``anchor'' the image-generation process~\cite{yao2021instance,garg2023instance}. However, such image generation is not relevant to the discriminative classification problem we aim to solve~\cite{bae2022noisy}. In addition, modeling $Z$ becomes troublesome with high-resolution inputs, leading to suboptimal reconstructions and transition-matrix estimation.
    Hence, we refrain from using $Z$ and define $p(\mathbf{x}| \mathbf{y})$ only over the finite set of training samples. This facilitates the direct optimization of $p(\mathbf{x}| \mathbf{y})$ and alleviates the difficulty in training an image generator~\cite{rolf2022resolving,wang2023aligning}. This is achieved by the following approximation:
    \begin{equation}
    p(\mathbf{x}|\mathbf{y}) \approx \frac{q(\mathbf{y} | \mathbf{x})}{\sum_{i=1}^{|\mathcal{D}|}q(\mathbf{y}| \mathbf{x}_{i})}. \label{eq:estimation_p_x_y}
    \end{equation}
    Here, the conditional probability $p(\mathbf{x}|\mathbf{y})$ is defined only for the available training samples $\mathbf{x}$ given their latent labels $\mathbf{y}$, so $p(\mathbf{x}|\mathbf{y})$ has dimension $|\mathcal{D}| \times |\mathcal{Y}|$. As noted in~\cite{rolf2022resolving}, $p(\mathbf{x}|\mathbf{y})$ is large when $q(\mathbf{y}| \mathbf{x})$ is also large relative to $q(\mathbf{y}| \mathbf{x}_{i})$ for all training samples $(\mathbf{x}_{i},\tilde{\mathbf{y}}_{i}) \in \mathcal{D}$.

    With $p(\mathbf{x}\mid \mathbf{y})$ approximated using the variational posterior $q(\mathbf{y}\mid \mathbf{x})$, we can obtain $p(\mathbf{x})$ via the following marginalization:
    \begin{equation}
        p(\mathbf{x}) = \sum_{\mathbf{y} \in \mathcal{Y}}p(\mathbf{x}|\mathbf{y})p(\mathbf{y}). \label{eq:estimation_p_x}
    \end{equation}
    
    Note that both approximations in Eq.~\eqref{eq:estimation_p_x_y} and Eq.~\eqref{eq:estimation_p_x} are dataset-level approximations that rely solely on the discriminative classifier $q(\mathbf{y}\mid \mathbf{x})$. They do not represent the true, intractable distributions.
    However, for classification tasks, these approximations enable much more efficient optimization than current generative methods~\cite{garg2023instance,yao2021instance} because they avoid training image generation models.

    \subsection{Approximation of $p(\mathbf{y})$ by partial label supervision}
    \label{sec:construct_partial}

    As depicted in Fig.~\ref{fig:causal_relationships}(a),
$Z$ and $Y$ jointly generate $X$. Based on the d-separation rule~\cite{pearl1988probabilistic}, $Z$ and $Y$ become dependent when $X$ is observed.
Traditional generative modeling assumes a uniform prior over $Y$ and constrains $X$ by estimating an informative $Z$. Because we remove $Z$ from our framework, $Y$ must be informative to better constrain the generation of $X$.

To approximate the clean-label prior $p(\mathbf{y})$, prior work~\cite{bae2022noisy,zhuang2023dygen} has explored multi-stage training, which is computationally intensive and time-consuming. Motivated by recent work on PLL decomposition~\cite{qiao2022decompositional}, we introduce Partial-Label Supervision (PLS) to construct an informative $p(\mathbf{y})$. Unlike traditional pseudo-labeling methods, which commit to a single, definitive prediction for a training sample with a noisy label and rely heavily on prediction confidence, our PLS constructs $p(\mathbf{y})$ as a set of candidate labels per training sample and the set is as small as possible while maintaining a high probability of covering the clean label $Y$.
Therefore, PLS is designed to balance two objectives: maximizing label coverage while minimizing label uncertainty, defined by:
    \begin{equation}
        \begin{split}
            \text{Coverage}&= \frac{1}{|\mathcal{D}|}\sum_{i=1}^{|\mathcal{D}|}\sum_{\mathbf{y} \in \mathcal{Y}}\mathbbm{1}\left[ \mathbf{y}_{i}= \mathbf{y}\right ] \times \mathbbm{1}\left[ p_{i}(\mathbf{y})>0 \right ], \\
            \text{Uncertainty}&= \frac{1}{|\mathcal{D}|}\sum_{i=1}^{|\mathcal{D}|}\sum_{\mathbf{y} \in \mathcal{Y}}\mathbbm{1}\left[p_{i}(\mathbf{y}) > 0\right],
        \end{split}
        \label{eq:metrics}
    \end{equation}
    where $\mathbf{y}_{i}$ is the latent clean label of the $i^{th}$ training sample, $p_{i}(\mathbf{y})$ denotes the prior label probability of the $i^{th}$ training sample for the class $\mathbf{y}\in \mathcal{Y}$, and $\mathbbm{1}(\cdot)$ is the indicator function.

    In Eq.~\eqref{eq:metrics}, coverage increases by pushing $p(\mathbf{y})$ toward a uniform distribution, whereas uncertainty is minimized when the latent clean label $\mathbf{y}_{i}$ is assigned the highest probability. These objectives are ultimately achieved when the prior $p_{i}(\mathbf{y})$ approaches a one-hot distribution concentrated on the clean label $\mathbf{y}_{i}$. To this end, we define the clean-label prior as:

    \begin{equation}
        p_{i}(\mathbf{y}(j)) = \frac{\tilde{\mathbf{y}}_{i}(j) + \mathbf{c}_{i}(j) + \mathbf{u}_{i}(j)}{Z},
        \label{eq:true_label_prior}
    \end{equation}
    where $j \in \{1,...,|\mathcal{Y}|\}$, $\tilde{\mathbf{y}}_{i}\in \mathcal{Y}$ is the noisy-label in the training set,
    $\mathbf{c}_{i}\in \Delta^{|\mathcal{Y}|}$
    denotes the candidate label built to increase coverage (
    $\Delta^{|\mathcal{Y}|}$
    denotes the probability simplex),
    $\mathbf{u}_{i}\in \Delta^{|\mathcal{Y}|}$
    represents the candidate label built to decrease uncertainty, and $Z$ is a normalization factor for ensuring $\sum_{\mathbf{y} \in \mathcal{Y}}p_{i}(\mathbf{y})=1$.
    We define $\mathbf{c}_{i}$ and $\mathbf{u}_{i}$ below.

    Since the variational posterior $q(\mathbf{y}|\mathbf{x})$ is optimized to recover likely candidates for the clean label,
    we maximize coverage by sampling from a moving average of $q(\mathbf{y}|\mathbf{x})$ for each training sample $\mathbf{x}_{i}$ at training iteration $t$ with:
    \begin{equation}
        \begin{split}
            \mathcal{C}_{i}^{(t)}&= \beta \times \mathcal{C}_{i}^{(t-1)}+ (1 - \beta) \times \bar{\mathbf{y}}_{i}^{(t)},
        \end{split}
        \label{eq:moving_average}
    \end{equation}
    where $\beta \in [0,1]$, and $\bar{\mathbf{y}}_{i}^{(t)}$ is the softmax output of $q(\mathbf{y}|\mathbf{x})$ that predicts the clean label given the input $\mathbf{x}_{i}$.
    For Eq.~\eqref{eq:moving_average}, $\mathcal{C}_{i}^{(t)}$ denotes the categorical distribution of the most likely labels for the $i^{th}$ training sample, which is used to sample the one-hot label $\mathbf{c}_{i}\sim \mathsf{Cat}(\mathcal{C}_{i}^{(t)})$.

    The minimization of uncertainty depends on our ability to detect clean-label and noisy-label samples. For clean samples, the label prior $p_{i}(\mathbf{y})$ should converge to a distribution focused on a few candidate labels. For noisy samples, this label prior should be close to uniform to retain a larger candidate set and reduce supervision confidence, resulting in slower fitting. To estimate the probability $w_{i}\in [0,1]$ that the training label is noisy,
    we compute the cross-entropy loss from the variational posterior $q(\mathbf{y}|\mathbf{x})$ output
    $\bar{\mathbf{y}}_{i}$ and the noisy-label $\tilde{\mathbf{y}}$ with:
    \begin{equation}
        \ell_{i}= - \tilde{\mathbf{y}}_{i}^{\top}\log \bar{\mathbf{y}}_{i}. \label{eq:loss_gmm}
    \end{equation}
    
    To obtain $w_i$, we model the distribution of sample-wise losses in Eq.~\eqref{eq:loss_gmm} using a two-component Gaussian Mixture Model (GMM)~\cite{bishop2006pattern}. The resulting $w_i$ corresponds to the posterior probability that the sample $\mathbf{x}_i$ originates from the noisy component, which is typically associated with higher losses.
    
    In contrast to the small-loss hypothesis explored in the existing literature~\cite{li2020dividemix,han2018co}, which associates noisy-label samples with the cross-entropy loss between the label $\tilde{\mathbf{y}}$ and the model output—often close to a one-hot distribution—our variational posterior model $q(\mathbf{y}|\mathbf{x})$ is trained to approximate the multiple PLS labels from $p(\mathbf{y})$. As a result, $q(\mathbf{y}|\mathbf{x})$ does not merely focus on minimizing loss; it seeks to capture the underlying label distribution and uncertainty.

\begin{figure}
        \centering
        \includegraphics[width=\linewidth]{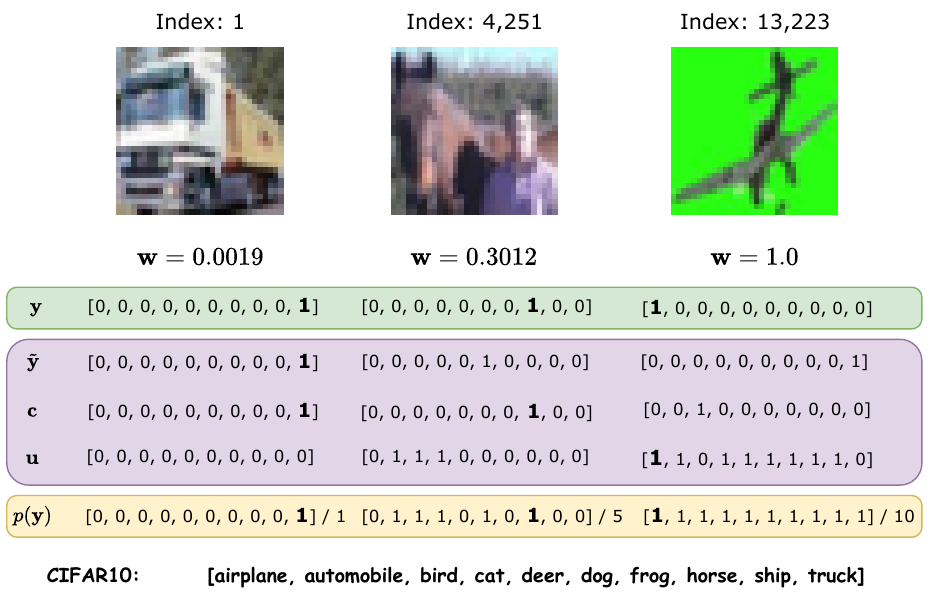}
        \caption{{Examples of CIFAR-10 images with 40\% IDN noise and PLS-constructed partial labels from Eq.~\eqref{eq:true_label_prior}. Here, $\mathbf{y}$ is the clean label; $\tilde{\mathbf{y}}$ is the noisy-label; $\mathbf{c}$ is the ``Coverage'' label from Eq.~\eqref{eq:moving_average}; $\mathbf{u}$ is the ``Uncertainty'' label from Eq.~\eqref{eq:min_ineff}; and $\mathbf{w}$ is the noisy-label probability from Eq.~\eqref{eq:loss_gmm}.}} 
        \label{fig:pls_sample}
    \end{figure}

    The label $\mathbf{u}_{i}$ is obtained by sampling from a uniform distribution over labels, where the number of sampled labels is proportional to the noise probability,
\begin{equation}
    \mathbf{u}_{i}\sim \mathcal{U}\left(\mathcal{Y}, \mathsf{round}(|\mathcal{Y}| \times w_{i}) \right), \label{eq:min_ineff}
\end{equation}
where $\mathsf{round}(|\mathcal{Y}| \times w_{i})$ denotes the number of labels drawn from the uniform distribution, rounded to the nearest integer. {Examples of PLS construction for each term of Eq.~\eqref{eq:true_label_prior} are shown in Fig.~\ref{fig:pls_sample}.}

    \subsection{Overall optimization process}
    \label{sec:training}

    The overall training and testing diagrams of our proposed framework are shown in Fig.~\ref{fig:architectures}, and the full training algorithm is presented in Alg.~\ref{alg:gnl}.

We now return to the optimization of Eq.~\eqref{eq:optim_goal} and Eq.~\eqref{eq:pyx_optim_goal}, where we define the neural network
$g_{\theta}:\mathcal{X}\to \Delta^{|\mathcal{Y}|}$
to represent the variational posterior $q(\mathbf{y}| \mathbf{x})$. This network outputs the categorical distribution for the clean label in the probability simplex
$\Delta^{|\mathcal{Y}|}$
given an image $\mathbf{x}\in \mathcal{X}$. The network
$f_{\phi}:\mathcal{X}\times \Delta^{|\mathcal{Y}|}\to \Delta^{|\mathcal{Y}|}$
denotes the instance-dependent label-transition module $p(\tilde{\mathbf{y}}| \mathbf{x}, \mathbf{y})$, which outputs the categorical distribution for the noisy training label given an image and the clean-label distribution from $g_{\theta}(\cdot)$.

The first term on the right-hand side (RHS) of Eq.~\eqref{eq:optim_goal} and Eq.~\eqref{eq:pyx_optim_goal} is optimized with the cross-entropy loss to maximize the expected log-likelihood (i.e., the M-step in the EM algorithm):
\begin{equation}
    \mathcal{L}_{CE}(\theta,\phi,\mathcal{D}) = \frac{1}{|\mathcal{D}| \times K}\sum_{(\mathbf{x}_i, \tilde{\mathbf{y}}_i) \in \mathcal{D}}\sum_{j = 1}^{K}\ell_{CE}(\tilde{\mathbf{y}}_{i}, f_{\phi}(\mathbf{x}_{i}, \hat{\mathbf{y}}_{i,j})),
    \label{eq:ce}
\end{equation}
where $\{\hat{\mathbf{y}}_{i,j}\}_{j=1}^{K}\sim \mathsf{Cat}(g_{\theta}(\mathbf{x}_{i}))$, with $\mathsf{Cat}(\cdot)$ denoting a categorical distribution from which $K$ independent label samples are drawn. The second term on the RHS of Eq.~\eqref{eq:optim_goal} and Eq.~\eqref{eq:pyx_optim_goal} uses the estimate of $p(\mathbf{x}|\mathbf{y})$ from Eq.~\eqref{eq:estimation_p_x_y} to minimize the $\mathsf{KL}[\cdot]$ divergence in the M-step, as in:
    \begin{equation}
        \begin{split}
            \mathcal{L}_{PRI}(\theta,\mathcal{D})&= \frac{1}{|\mathcal{D}|}\times \\
            &\sum_{(\mathbf{x}_i, \tilde{\mathbf{y}}_i) \in \mathcal{D}}\mathsf{KL}\left[ g_{\theta}(\mathbf{x}_{i}) \Big \| c_{i}\times \frac{g_{\theta}(\mathbf{x}_{i})}{\sum_{j}g_{\theta}(\mathbf{x}_{j})}\odot \mathbf{p}_{i}\right],
        \end{split}
        \label{eq:partial_loss}
    \end{equation}
    where $\mathbf{p}_{i}= [p_{i}(j=1),...,p_{i}(j=|\mathcal{Y}|)] \in
    \Delta^{|\mathcal{Y}|}$
    is the clean-label prior defined in Eq.~\eqref{eq:true_label_prior}, $c_{i}$ is a normalization factor, and $\odot$ denotes element-wise multiplication.
    The optimization of the last term on the RHS of Eq.~\eqref{eq:optim_goal} and Eq.~\eqref{eq:pyx_optim_goal} represents the E-step in the EM derivation~\cite{dempster1977maximum,kingma2017variational}.
    This E-step consists of minimizing the $\mathsf{KL}[\cdot]$ divergence with:
    \begin{equation}
        \begin{split}
            \mathcal{L}_{KL}(\theta,\phi,&\mathcal{D}) = \frac{1}{|\mathcal{D}|}\times \\
            &\sum_{(\mathbf{x}_i, \tilde{\mathbf{y}}_i) \in \mathcal{D}}\mathsf{KL}\left[ g_{\theta}(\mathbf{x}_{i}) \Big \| f_{\phi}(\mathbf{x}_{i}, g_{\theta}(\mathbf{x}_{i})) \odot \mathbf{p}_{i}\right]
        \end{split}
        \label{eq:KL_divergence_pxy}
    \end{equation}
    for $\log p(\mathbf{x}| \mathbf{y})$ described in Sec.~\ref{sec:framework}, 
    and
    \begin{equation}
        \begin{split}
            &\mathcal{L}_{KL}(\theta,\phi, \mathcal{D}) = \frac{1}{|\mathcal{D}|}\times \\
            &\sum_{(\mathbf{x}_i, \tilde{\mathbf{y}}_i) \in \mathcal{D}}\mathsf{KL}\left[ g_{\theta}(\mathbf{x}_{i}) \Big \| f_{\phi}(\mathbf{x}_{i}, g_{\theta}(\mathbf{x}_{i})) \odot \sum_{y=k}^{|\mathcal{Y}|}{\frac{g_{\theta}(\mathbf{x}_{i})}{\sum_{j}g_{\theta}(\mathbf{x}_{j})} \mathbf{p}_i }\right]
        \end{split}
        \label{eq:KL_divergence_pyx}
    \end{equation}
    for $\log p(\mathbf{y}| \mathbf{x})$ elaborated in Sec.~\ref{sec:pyx_optim}. 

    Hence, the complete loss function to minimize is:
    \begin{equation}
        \mathcal{L}(\theta,\phi,\mathcal{D}) = \mathcal{L}_{CE}(\theta,\phi,\mathcal{D})+ \lambda_1 \mathcal{L}_{PRI}(\theta,\mathcal{D}) + \lambda_2 \mathcal{L}_{KL}(\theta,\phi,\mathcal{D}), \label{eq:final}
    \end{equation}
    where $\lambda_1, \lambda_2$ are loss weights for $ \mathcal{L}_{PRI}$ and $\mathcal{L}_{KL}$, respectively.

    During testing, a test image $\mathbf{x}$ is associated with a class categorical distribution produced by $g_{\theta}(\mathbf{x})$.

    \begin{algorithm}
        [t]
        
        \caption{EM training with Partial Label Supervision (PLS).}
        \label{alg:gnl}
        \begin{algorithmic}
            [1] \STATE \textbf{Input:} $\mathcal{D}=\{(\mathbf{x}_{i},\tilde{\mathbf{y}}_{i})\}_{i=1}^{|\mathcal{D}|}$, number of epochs $T$, batch size $B$ \STATE \textbf{Output:} Classifier $g_{\theta}$ \STATE Initialize $\theta, \phi$ and moving average $\mathcal{C}_{i}^{(0)}$ for $i=1...|\mathcal{D}|$ \FOR{epoch $t=1$ to $T$} \FOR{mini-batch $\{(\mathbf{x}_{i},\tilde{\mathbf{y}}_{i})\}_{i=1}^{B}$} \STATE $\bar{\mathbf{y}}_{i}\leftarrow g_{\theta}(\mathbf{x}_{i})$ \STATE \textbf{Construct Prior} $p_{i}(\mathbf{y})$ (Sec.~\ref{sec:construct_partial}): \STATE \quad Coverage: update $\mathcal{C}_{i}^{(t)}$ (Eq.~\eqref{eq:moving_average}), sample $\mathbf{c}_{i}$ \STATE \quad Uncertainty: obtain $\mathbf{u}_{i}$ (Eq.~\eqref{eq:min_ineff}) using loss from Eq.~\eqref{eq:loss_gmm} \STATE \quad Combine: $p_{i}(\mathbf{y}) \leftarrow (\tilde{\mathbf{y}}_{i}+ \mathbf{c}_{i}+ \mathbf{u}_{i})/Z$ (Eq.~\eqref{eq:true_label_prior})  \STATE \textbf{E-step} (Minimize KL for a given causal process): \IF{$Y \rightarrow X$ (causal)} \STATE Compute $\mathcal{L}_{KL}$ via Eq.~\eqref{eq:KL_divergence_pxy} \ELSE[$X \rightarrow Y$ (anti-causal)] \STATE Compute $\mathcal{L}_{KL}$ via Eq.~\eqref{eq:KL_divergence_pyx} \ENDIF \STATE \textbf{M-step} (Maximize ELBO): \STATE \quad Compute $\mathcal{L}_{CE}$ (Eq.~\eqref{eq:ce}) and $\mathcal{L}_{PRI}$ (Eq.~\eqref{eq:reverse_partial_loss})
            \STATE \textbf{Update:} Minimize $\mathcal{L}=\mathcal{L}_{CE}+\mathcal{L}_{PRI}+\mathcal{L}_{KL}$ (Eq.~\eqref{eq:final}) w.r.t. $\theta, \phi$ \ENDFOR \ENDFOR
        \end{algorithmic}
    \end{algorithm}

\newtheorem{lemma}{Lemma}
\newtheorem{corollary}{Corollary}
\newtheorem{remark}{Remark}
\newtheorem{proposition}{Proposition}

\subsection{Why partial labels yield better pseudo-labels}
\label{sec:theory-partial-labels}

We begin by defining a measure of pseudo-label quality: the expected squared $\ell_{2}$-norm of the error between the true label and the pseudo-label, a quantity widely used in ensemble learning~\cite{ueda1996generalization,krogh1994neural,wood2023unified}. Let $p_{c}= P(\hat{\mathbf{y}}= \mathbf{y})$ denote the probability that the one-hot predicted label $\hat{\mathbf{y}}$ matches the one-hot clean label $\mathbf{y}$.

\begin{lemma}[Hard re-labeling error]
\label{lem:hard}
\[
\mathbb{E}\!\left[\|\mathbf{y}-\hat{\mathbf{y}}\|^2\right] \;=\; 2(1-p_c).
\]
\end{lemma}

\begin{proof}
\renewcommand{\qedsymbol}{}
The expected squared error for standard relabeling is:
\begin{enumerate}
    \item With probability $p_{c}$, the prediction is correct (i.e., $\hat{\mathbf{y}}= \mathbf{y}$), so the error vector is $\mathbf{0}$ and its squared norm is $0$.
    \item With probability $1 - p_{c}$, the prediction is incorrect, hence the error vector has one element equal to $1$ (at the true class index) and one element equal to $-1$ (at the incorrectly predicted index). The squared $\ell_{2}$-norm is $1^{2}+ (-1)^{2}= 2$.
\end{enumerate}
            Therefore,
            \[
                \mathbb{E}\!\left[ \|\mathbf{y}- \hat{\mathbf{y}}\|^{2}\right] = 2(1 - p_{c}).
            \]
\end{proof}

\begin{lemma}[Uniform partial-label error]
\label{lem:partial}
\[
\mathbb{E}\!\left[\|\mathbf{y}-\bar{\mathbf{y}}\|^2\right] \;=\; 1 + \frac{1-2p_k}{K}.
\]
\end{lemma}

\begin{proof}
\renewcommand{\qedsymbol}{}
We consider a partial label $\bar{\mathbf{y}}$ that is uniform over a candidate set of size $K$. Let $p_{k}$ be the probability that the true label is included in the candidate set; then we have two cases:
\begin{enumerate}
    \item If the clean label is present in $\bar{\mathbf{y}}$, the clean label has probability
    $\bar{y}_{y}= 1/K$. Then $\|\bar{\mathbf{y}}\|_{2}^{2}= \sum_{i=1}^{\|Y\|}\bar{y}_{i}^{2}= K \cdot (1/K^{2}) = 1/K$, and
        \begin{equation*}
        \scalebox{0.9}{$
            \begin{split}
                \|\mathbf{y}- \bar{\mathbf{y}}\|^{2}&= (1 - \bar{y}_{y})^{2}+ \sum_{j \neq y}\bar{y}_{j}^{2} = 1 - \frac{2}{K}+ \frac{1}{K^{2}}+ \frac{K-1}{K^{2}}\\&= 1 - \frac{1}{K}.
            \end{split}
            $}
        \end{equation*}

                \item If the true label is not included, $\bar{y}_{y}= 0$ and $\|\bar{\mathbf{y}}\|_{2}^{2}= 1/K$, so
                    \[
                        \|\mathbf{y}- \bar{\mathbf{y}}\|^{2}= 1 + \frac{1}{K}.
                    \]
            \end{enumerate}
            Hence,
            \[\scalebox{0.95}{$
                \mathbb{E}\!\left[ \|\mathbf{y}- \bar{\mathbf{y}}\|^{2}\right] = p_{k}\!\left(1 - \frac{1}{K}\right) + (1 - p_{k}) \!\left(1 + \frac{1}{K}\right) = 1 + \frac{1 - 2p_{k}}{K}.
                $}
            \]
\end{proof}

We analyze two bounds for different noisy-label scenarios. In the best case, the model confidently predicts the correct label ($p_{c}=1$) and the partial label has $p_{k}=1$ with $K=2$ (including the noisy label and the model prediction). Then the hard relabeling error is $0$, and the partial-label error is $1 - \tfrac{1}{K}= 0.5$, so hard relabeling performs better. 
However, in the worst case, when the model provides no useful information ($p_{c}= 0$), or the candidate set excludes the true label ($p_{k}= 0$) with $K = \|Y\| - 1$, the expected errors are $2$ (relabeling) versus $1 + \tfrac{1}{K}$ (partial-label), which approaches $1$ when $K$ is large. This shows that partial labels provide a more robust learning signal when model predictions are unreliable.

\begin{proposition}[Sensitivity of the partial-label error]
\label{prop:sensitivity}
For $\,\mathbb{E}\!\left[\|\mathbf{y}-\bar{\mathbf{y}}\|^2\right]=1+\frac{1-2p_k}{K}\,$, we have:
\[
\frac{\partial}{\partial p_k}\,\mathbb{E}\!\left[\|\mathbf{y}-\bar{\mathbf{y}}\|^2\right] = -\frac{2}{K},
\qquad
\frac{\partial}{\partial K}\,\mathbb{E}\!\left[\|\mathbf{y}-\bar{\mathbf{y}}\|^2\right] = \frac{2p_k-1}{K^2}.
\]
Thus, for any fixed $K$, increasing coverage ($p_k$) always reduces the expected error at a constant rate of $2/K$.
The effect of enlarging the candidate set depends on coverage: if $p_k<0.5$, increasing $K$ decreases the error; if $p_k>0.5$, increasing $K$ increases the error; and if $p_k=0.5$, changing $K$ has no first-order effect.
\end{proposition}

\begin{remark}[Design implication]
The analysis suggests a two-stage strategy: first \emph{maximize coverage} $p_k$ (e.g., via ensembling/EMA and stochastic sampling to reduce confirmation bias); once coverage is high, \emph{shrink $K$} using uncertainty-based filtering to further reduce the error.
This aligns with our ablations in Tab.~\ref{tab:abl_grouped}.

\end{remark}

    \section{Experiments}

    We report results on synthetic benchmarks using CIFAR-10 and CIFAR-100~\cite{krizhevsky2009learning} with instance-dependent noise, and on the NLP datasets AG News~\cite{zhang2015character} and 20~Newsgroups~\cite{lang1995newsweeder} with symmetric, asymmetric, and instance-dependent noise. We also evaluate on real-world datasets, including CIFAR-10N and CIFAR-100N~\cite{wei2021learning}, Animal-10N~\cite{song2019selfie}, Red Mini-ImageNet~\cite{jiang2017mentornet}, Clothing1M~\cite{xiao2015learning}, and Mini-WebVision~\cite{li2017webvision}.

    \begin{table*}
        []
        \centering
        \scalebox{0.87}{
        \begin{tabular}{ >{\columncolor[HTML]{EFEFEF}}l |cccccc|cc}
            \hline
                                   & \cellcolor[HTML]{EFEFEF}IDN CIFAR10/100                   & \cellcolor[HTML]{EFEFEF}CIFAR10/100 N        & \cellcolor[HTML]{EFEFEF}Red Mini-ImageNet & \cellcolor[HTML]{EFEFEF}Animal-10N & \cellcolor[HTML]{EFEFEF}Clothing1M & \cellcolor[HTML]{EFEFEF}Mini-WebVision & \cellcolor[HTML]{EFEFEF}AG NEWS                 & \cellcolor[HTML]{EFEFEF}20newsgroup \\
            \hline
            Baseline reference     & kMEIDTM~\cite{cheng2022instance}                          & Real-world ~\cite{DBLP:conf/iclr/WeiZ0L0022} & FaMUS~\cite{xu2021faster}                 & Nested~\cite{chen2021boosting}     & CausalNL~\cite{yao2021instance}    & UNICON~\cite{karim2022unicon}          & \multicolumn{2}{c}{DyGEN~\cite{zhuang2023dygen}} \\
            Backbone               & ResNet-34                                                 & ResNet-34                                    & Pre-act ResNet-18                         & VGG-19BN                           & ResNet-50$^{1}$                    & InceptionNetV2                         & \multicolumn{2}{c}{BERT$^{2}$}                   \\
            Training epochs        & 150                                                       & 120                                          & 150                                       & 100                                & 80                                 & 100                                    & \multicolumn{2}{c}{10}                           \\
            Batch size             & 128                                                       & 128                                          & 128                                       & 128                                & 64                                 & 64                                     & \multicolumn{2}{c}{64}                           \\
            Learning rate          & 0.02                                                      & 0.02                                         & 0.02                                      & 0.02                               & 2e-3                               & 0.01                                   & \multicolumn{2}{c}{1e-4}                         \\
            Optimizer/Weight decay & SGD / 5e-4                                                & SGD / 5e-4                                   & SGD / 5e-4                                & SGD / 1e-3                         & SGD / 1e-3                         & SGD / 1e-4                             & \multicolumn{2}{c}{Adam / --}                    \\
            LR decay at epochs     & 0.1 / 100                                                 & 0.1 / 80                                     & 0.1 / 100                                 & 0.1 / 50                           & 0.1 / 40                           & 0.1 / 80                               & \multicolumn{2}{c}{Linear decay}                 \\
            Data augmentation      & \multicolumn{6}{c|}{Random Crop / Random horizontal flip} & \multicolumn{2}{c}{--}                        \\
            $\lambda_1 / \lambda_2$ & \multicolumn{6}{c|}{1.0 / 1.0}                            & \multicolumn{2}{c}{1.0 / 1.0}                   \\
            $\beta$                & \multicolumn{6}{c|}{0.9}                                  & \multicolumn{2}{c}{0.9}                       \\
            $K$                    & \multicolumn{6}{c|}{1}                                    & \multicolumn{2}{c}{1}                         \\
            \hline
        \end{tabular}}
        \caption{Hyper-parameter setups and baseline references for all experimental datasets. $^{1}$ means ImageNet pre-train. $^{2}$ means BERT$_{base}$~\cite{devlin2018bert} pre-train.}
        \label{tab:implementation_detail}
    \end{table*}
    \label{sec:datasets}

    The \textbf{CIFAR-10/100} datasets comprise 50K training images and 10K test images with a resolution of $32\times 32\times 3$, where CIFAR-10 has 10 classes and CIFAR-100 has 100 classes. We follow prior work~\cite{xia2020part} and synthetically generate instance-dependent noise (IDN) at rates \{0.2, 0.3, 0.4, 0.5\}. The \textbf{20~Newsgroups} dataset contains 9{,}051 training samples and 7{,}532 test samples across 20 classes. \textbf{AG~News} contains 40K training samples and 7{,}600 test samples across 4 classes. Following DyGEN~\cite{zhuang2023dygen}, we synthetically generate symmetric, asymmetric, and instance-dependent noise at rates \{0.2, 0.4\}. The \textbf{CIFAR-10N/CIFAR-100N} datasets provide real-world annotations for the original CIFAR-10/100 images; we evaluate the \{aggre, random1, random2, random3, worse\} splits for CIFAR-10N and the \{noisy\} split for CIFAR-100N.
    \textbf{Red Mini-ImageNet} is a real-world dataset with images annotated using the Google Cloud Data Labeling Service. It has 100 classes, each containing 600 images from ImageNet; images are resized to $32 \times 32$ pixels from the original $84 \times 84$ pixels to enable a fair comparison with other baselines~\cite{xu2021faster}. \textbf{Animal-10N} is a real-world dataset containing 10 animal species with five pairs of similar-looking classes (wolf and coyote, hamster and guinea pig, etc.). The training set has 50K images and the test set has 10K images; we follow the same setup as in~\cite{chen2021boosting}. \textbf{Clothing1M} is a real-world dataset with 100K images and 14 classes. The labels are automatically generated from surrounding text, with an estimated noise rate of 38.5\%. The dataset also contains clean training, clean validation, and clean test sets with 50K, 14K, and 10K images, respectively; we do not use the clean training and validation sets. The clean test set is used only to measure model performance. \textbf{Mini-WebVision} is a real-world dataset that contains 2.4 million images collected from the web. Following prior work~\cite{karim2022unicon}, we compare baselines on the first 50 classes and evaluate on both the WebVision and ImageNet test sets.

    \subsection{Causal relationships in each dataset} 
    
    We begin by analyzing the causal structure of each dataset:
\begin{itemize}
    \item CIFAR-10~\cite{krizhevsky2009learning} was collected based on predefined classes (a subset of the Tiny Images dataset), aligning with the causal relationship $Y \rightarrow X$ in~\cite{yao2023better}.
    
    \item 20~Newsgroups~\cite{lang1995newsweeder} is harvested from existing Usenet posts and labeled by source newsgroup, aligning with the causal relationship $Y \rightarrow X$ from~\cite{jin2021causal}.
        
    \item AG~News~\cite{zhang2015character} is scraped from news outlets and labeled using source/topic metadata, aligning with the anti-causal relationship $X \rightarrow Y$ from~\cite{jin2021causal}.
    
    \item Red Mini-ImageNet is derived from ImageNet~\cite{deng2009imagenet}. It follows a different generation process from CIFAR-10, in which images are web-crawled first and then annotated by humans. This implies that the data distribution is more diverse and complex than the label distribution, and therefore $X \rightarrow Y$.
    
    \item Animal-10N~\cite{song2019selfie} contains real-world animal images collected for ten predefined classes with human label noise; this aligns with the causal relationship $Y \rightarrow X$ from~\cite{yao2023better}.
    
    \item Clothing1M~\cite{xiao2015learning} crawls images from shopping sites and assigns labels using surrounding text for the given 14 categories; this aligns with the causal relationship $Y \rightarrow X$ from~\cite{yao2023better}.
    
    \item Mini-WebVision~\cite{li2017webvision} is a subset of WebVision built by querying web sources and inheriting noisy labels/metadata, often containing OOD samples; this aligns with the anti-causal relationship $X \rightarrow Y$ in~\cite{yao2023better}.
\end{itemize}

    \subsection{Implementation details}
\label{sec:implementation}  In Table~\ref{tab:implementation_detail},
we describe the implementation details of our method for each dataset, including the main reference for each corresponding baseline. In addition, for Clothing1M, we sample 1{,}000 mini-batches from the training set; in each mini-batch, we ensure that all 14 classes are evenly sampled to form a pseudo-balanced learning setting. For Clothing1M, we first resize images to $256 \times 256$ and then randomly crop to $224 \times 224$ with random horizontal flipping. The number of warm-up epochs is 10 for CIFAR and Red Mini-ImageNet, and 1 for Clothing1M.

Note that the approximation of the generative model in Eq.~\eqref{eq:estimation_p_x_y} is performed within each mini-batch, not over the entire dataset. 
Following the discussion by Rolf et al.~\cite{rolf2022resolving}, the minimization of $\mathcal{L}_{PRI}(\cdot)$ can be done via the reverse KL:
\begin{equation}
    \begin{split}
        \mathcal{L}_{PRI\_R}(\theta,\mathcal{D})&= \frac{1}{|\mathcal{D}|}\times \\
        &\sum_{(\mathbf{x}_i, \tilde{\mathbf{y}}_i) \in \mathcal{D}}\mathsf{KL}\left[ c_{i}\times \frac{ g_{\theta}(\mathbf{x}_{i})}{\sum_{j}g_{\theta}(\mathbf{x}_{j})}\odot \mathbf{p}_{i}\Big \| g_{\theta}(\mathbf{x}_{i}) \right].
    \end{split}
    \label{eq:reverse_partial_loss}
\end{equation}
This reverse KL divergence also encourages the model and the implied posterior to be close. In fact, the KL and reverse KL losses are equivalent when $\sum_{j}g_{\theta}(\mathbf{x}_{j})$ has a uniform distribution over the classes in $\mathcal{Y}$ and the prior $p_{i}(\mathbf{y})$ is uniform. We tried both versions of this loss (i.e., Eq.~\eqref{eq:partial_loss} and Eq.~\eqref{eq:reverse_partial_loss}), with the reverse KL generally producing better results, as shown in the ablation study in Sec.~\ref{sec:loss_fn_analysis}. For this reason, all experiments in Sec.~\ref{sec:experimental_results} use the reverse KL loss.

In most experiments, we evaluate our approach under both optimization goals, $p(X|Y)$ and $p(Y|X)$, following Eq.~\eqref{eq:KL_divergence_pxy} and Eq.~\eqref{eq:KL_divergence_pyx}, respectively. We denote these approaches as ``Ours - $p(X|Y)$'' and ``Ours - $p(Y|X)$''. For the real-world datasets Animal-10N, Red Mini-ImageNet, and Clothing1M, we also evaluate an ensemble of two networks. Our code is implemented in PyTorch, and experiments are performed on NVIDIA RTX~3090 GPUs.

    \begin{table*}
        []
        \centering
        \begin{tabular}{ >{\columncolor[HTML]{EFEFEF}}l |cccc|cccc}
            \hline
            \cellcolor[HTML]{EFEFEF}                                          & \multicolumn{4}{c|}{\cellcolor[HTML]{EFEFEF}CIFAR-10} & \multicolumn{4}{c}{\cellcolor[HTML]{EFEFEF}CIFAR-100} \\
            \cline{2-9} \multirow{-2}{*}{\cellcolor[HTML]{EFEFEF}Method}      & $20\%$                                                & $30\%$                                               & $40\%$                    & $50\%$                    & $20\%$                    & $30\%$                    & $40\%$                    & $50\%$                    \\
            \hline
            CE                                                                & 86.93 $\pm$ 0.17                                      & 82.42 $\pm$ 0.44                                     & 76.68 $\pm$ 0.23          & 58.93 $\pm$ 1.54          & 63.94 $\pm$ 0.51          & 61.97 $\pm$ 1.16          & 58.70 $\pm$ 0.56          & 56.63 $\pm$ 0.69          \\
            DMI~\cite{Xu_NeurIPS_2019_Information_Theoretic_Mutual_Info_Loss} & 89.99 $\pm$ 0.15                                      & 86.87 $\pm$ 0.34                                     & 80.74 $\pm$ 0.44          & 63.92 $\pm$ 3.92          & 64.72 $\pm$ 0.64          & 62.80 $\pm$ 1.46           & 60.24 $\pm$ 0.63          & 56.52 $\pm$ 1.18          \\
            Forward~\cite{patrini2017making}                                  & 89.62 $\pm$ 0.14                                      & 86.93 $\pm$ 0.15                                     & 80.29 $\pm$ 0.27          & 65.91 $\pm$ 1.22          & 67.23 $\pm$ 0.29          & 65.42 $\pm$ 0.63          & 62.18 $\pm$ 0.26          & 58.61 $\pm$ 0.44          \\
            CoTeaching~\cite{han2018co}                                       & 88.43 $\pm$ 0.08                                      & 86.40 $\pm$ 0.41                                     & 80.85 $\pm$ 0.97          & 62.63 $\pm$ 1.51          & 67.40 $\pm$ 0.44          & 64.13 $\pm$ 0.43          & 59.98 $\pm$ 0.28          & 57.48 $\pm$ 0.74          \\
            TMDNN~\cite{DBLP:conf/icml/YangYHLXNL22}                          & 88.14 $\pm$ 0.66                                      & 84.55 $\pm$ 0.48                                     & 79.71 $\pm$ 0.95          & 63.33 $\pm$ 2.75          & 66.62 $\pm$ 0.85          & 64.72 $\pm$ 0.64          & 59.38 $\pm$ 0.65          & 55.68 $\pm$ 1.43          \\
            PartT~\cite{xia2020part}                                          & 89.33 $\pm$ 0.70                                      & 85.33 $\pm$ 1.86                                     & 80.59 $\pm$ 0.41          & 64.58 $\pm$ 2.86          & 65.33 $\pm$ 0.59          & 64.56 $\pm$ 1.55          & 59.73 $\pm$ 0.76          & 56.80 $\pm$ 1.32          \\
            kMEIDTM~\cite{cheng2022instance}                                  & 92.26 $\pm$ 0.25                                      & 90.73 $\pm$ 0.34                                     & 85.94 $\pm$ 0.92          & 73.77 $\pm$ 0.82          & 69.16 $\pm$ 0.16          & 66.76 $\pm$ 0.30          & 63.46 $\pm$ 0.48          & 59.18 $\pm$ 0.16          \\
            \hline
            CausalNL~\cite{yao2021instance}                                   & 81.47 $\pm$ 0.32                                      & 80.38 $\pm$ 0.44                                     & 77.53 $\pm$ 0.45          & 67.39 $\pm$ 1.24          & 41.47 $\pm$ 0.43          & 40.98 $\pm$ 0.62          & 34.02 $\pm$ 0.95          & 32.13 $\pm$ 2.23          \\
            Ours - $p(Y|X)$                                                   & \textbf{92.63 $\pm$ 0.23}                             & 91.39 $\pm$ 0.31                                     & 90.91 $\pm$ 0.43          & 89.77 $\pm$ 0.54          & 70.73 $\pm$ 0.47          & 68.46 $\pm$ 1.03          & 66.21 $\pm$ 0.42          & 61.07 $\pm$ 0.13          \\
            Ours - $p(X|Y)$                                                   & 92.60 $\pm$ 0.44                                      & \textbf{91.96 $\pm$ 0.23}                            & \textbf{91.11 $\pm$ 0.08} & \textbf{90.57 $\pm$ 0.18} & \textbf{72.51 $\pm$ 0.39} & \textbf{70.45 $\pm$ 0.34} & \textbf{68.56 $\pm$ 0.36} & \textbf{64.00 $\pm$ 0.89} \\
            \hline
        \end{tabular}
        \caption{Accuracy (\%) on the test set for IDN problems on CIFAR-10/100. Most results are from kMEIDTM~\cite{cheng2022instance}. Experiments are repeated five times to compute mean $\pm$ standard deviation. The top part shows discriminative and the bottom part shows generative models. Best results are highlighted.}
        \label{tab:idn_cifar}
    \end{table*}
    
    \begin{table*}
        []
        \centering
        \scalebox{0.85}{
        \begin{tabular}{ >{\columncolor[HTML]{EFEFEF}}l |cccccc|cccccc}
            \hline
            Dataset                         & \multicolumn{6}{c|}{\cellcolor[HTML]{EFEFEF}AG News} & \multicolumn{6}{c}{\cellcolor[HTML]{EFEFEF}20~Newsgroups} \\
            \hline
            Method/Noise                    & 20\% SN                                              & 40\% SN                                                  & 20\% ASN       & 40\% ASN       & 20\% IDN       & 40\% IDN       & 20\% SN        & 40\% SN        & 20\% ASN       & 40\% ASN       & 20\% IDN       & 40\% IDN       \\
            \hline
            Base                            & 82.08                                                & 78.17                                                    & 81.43          & 77.15          & 85.59          & 75.86          & 78.84          & 70.81          & 74.44          & 56.18          & 77.38          & 69.81          \\
            Co-Teaching~\cite{han2018co}    & 82.99                                                & 78.79                                                    & 81.96          & 78.07          & 87.85          & 76.52          & 77.66          & 69.25          & 77.51          & 67.26          & 77.45          & 73.76          \\
            JoCoR~\cite{wei2020combating}   & 83.82                                                & 81.26                                                    & 85.88          & 77.98          & 87.04          & 79.93          & 80.92          & 73.27          & 81.01          & 69.40          & 81.57          & 74.19          \\
            CR~\cite{zhou2021learning}      & 89.10                                                & 78.40                                                    & 89.03          & 74.52          & 87.48          & 75.29          & 81.61          & 74.33          & 80.62          & 67.63          & 82.58          & 76.33          \\
            DualT~\cite{dualT2020nips}      & 83.66                                                & 80.84                                                    & 82.11          & 79.03          & 86.47          & 78.84          & 78.92          & 73.39          & 74.66          & 67.82          & 77.16          & 70.61          \\
            CausalNL~\cite{yao2021instance} & 86.44                                                & 82.74                                                    & 89.87          & 79.80          & 89.00          & 84.62          & 81.08          & 74.43          & 81.22          & 71.25          & 82.57          & 78.91          \\
            NPC~\cite{bae2022noisy}         & 82.83                                                & 75.04                                                    & 83.94          & 77.69          & 86.28          & 77.38          & 79.82          & 72.96          & 78.88          & 61.69          & 79.97          & 75.19          \\
            Ours - $p(Y|X)$                 & 89.86                                                & 89.60                                                    & 88.97          & 87.05          & 89.27          & 87.47          & 81.61          & 76.47          & 80.19          & 75.02          & 83.16          & 80.74          \\
            Ours - $p(X|Y)$                 & \textbf{90.72}                                       & \textbf{90.35}                                           & \textbf{90.54} & \textbf{89.80} & \textbf{90.05} & \textbf{88.77} & \textbf{82.63} & \textbf{78.91} & \textbf{82.71} & \textbf{81.63} & \textbf{83.53} & \textbf{81.04} \\
            \hline
            CR w/ NPC                       & 89.69                                                & 83.21                                                    & 89.01          & 82.54          & 88.25          & 86.41          & 83.09          & 77.96          & 83.13          & 73.50          & 83.47          & 80.47          \\
            DyGEN w/ NPC                    & 91.42                                                & 89.80                                                    & 91.37          & 90.43          & 91.41          & 88.90          & 83.82          & 79.56          & 83.63          & 81.98          & 84.07          & 81.54          \\
            Ours - $p(Y|X)$ w/ NPC          & 91.28                                                & 90.27                                                    & 91.81          & 90.81          & 91.26          & \textbf{89.19} & 83.73          & 79.24          & 81.38          & 82.11          & 83.53          & 81.47          \\
            Ours - $p(X|Y)$ w/ NPC          & \textbf{92.34}                                       & \textbf{91.63}                                           & \textbf{91.87} & \textbf{91.37} & \textbf{91.78} & 89.15          & \textbf{85.15} & \textbf{81.07} & \textbf{84.65} & \textbf{84.18} & \textbf{84.58} & \textbf{82.34} \\
            \hline
        \end{tabular}}
        \caption{Accuracy (\%) on news topic classification benchmarks with different label-noise types (SN = symmetric noise, ASN = asymmetric noise, IDN = instance-dependent noise) and rates. All baselines are taken from DyGEN~\cite{zhuang2023dygen}. In both tables, the bottom part shows methods that use NPC as a post-processing method. Best results are highlighted in each part.}
        \label{tab:nlp_news}
    \end{table*}

    \begin{table*}
        []
        \centering
        \begin{tabular}{ >{\columncolor[HTML]{EFEFEF}}l |ccccc|c}
            \hline
            \cellcolor[HTML]{EFEFEF}                                         & \multicolumn{5}{c|}{\cellcolor[HTML]{EFEFEF}CIFAR-10N} & \cellcolor[HTML]{EFEFEF}CIFAR-100N \\
            \cline{2-7} \multirow{-2}{*}{\cellcolor[HTML]{EFEFEF}Method}     & Aggregate                                              & Random 1                          & Random 2                  & Random 3                  & Worst                     & Noisy                     \\
            \hline
            CE                                                               & 87.77 $\pm$ 0.38                                       & 85.02 $\pm$ 0.65                  & 86.46 $\pm$ 1.79          & 85.16 $\pm$ 0.61          & 77.69 $\pm$ 1.55          & 55.50 $\pm$ 0.66          \\
            Forward T~\cite{patrini2017making}                               & 88.24 $\pm$ 0.22                                       & 86.88 $\pm$ 0.50                  & 86.14 $\pm$ 0.24          & 87.04 $\pm$ 0.35          & 79.79 $\pm$ 0.46          & 57.01 $\pm$ 1.03          \\
            T-Revision~\cite{xia2019anchor}                                  & 88.52 $\pm$ 0.17                                       & 88.33 $\pm$ 0.32                  & 87.71 $\pm$ 1.02          & 80.48 $\pm$ 1.20          & 80.48 $\pm$ 1.20          & 51.55 $\pm$ 0.31          \\
            Peer Loss \cite{liu2020peer}                                     & 90.75 $\pm$ 0.25                                       & 89.06 $\pm$ 0.11                  & 88.76 $\pm$ 0.19          & 88.57 $\pm$ 0.09          & 82.00 $\pm$ 0.60          & 57.59 $\pm$ 0.61          \\
            Positive-LS~\cite{Lukasik_ICML_2020_label_smoothing_label_noisy} & 91.57 $\pm$ 0.07                                       & 89.80 $\pm$ 0.28                  & 89.35 $\pm$ 0.33          & 89.82 $\pm$ 0.14          & 82.76 $\pm$ 0.53          & 55.84 $\pm$ 0.48          \\
            F-Div~\cite{Wei_ICLR_2021_f_Divergence}                          & 91.64 $\pm$ 0.34                                       & 89.70 $\pm$ 0.40                  & 89.79 $\pm$ 0.12          & 89.55 $\pm$ 0.49          & 82.53 $\pm$ 0.52          & 57.10 $\pm$ 0.65          \\
            Negative-LS~\cite{wei2021understanding}                          & 91.97 $\pm$ 0.46                                       & 90.29 $\pm$ 0.32                  & 90.37 $\pm$ 0.12          & 90.13 $\pm$ 0.19          & 82.99 $\pm$ 0.36          & 58.59 $\pm$ 0.98          \\
            CORES$^{2}$~\cite{cheng2020learning}                             & 91.23 $\pm$ 0.11                                       & 89.66 $\pm$ 0.32                  & 89.91 $\pm$ 0.45          & 89.79 $\pm$ 0.50          & 83.60 $\pm$ 0.53          & 61.15 $\pm$ 0.73          \\
            VolMinNet~\cite{li2021provably}                                  & 89.70 $\pm$ 0.21                                       & 88.30 $\pm$ 0.12                  & 88.27 $\pm$ 0.09          & 88.19 $\pm$ 0.41          & 80.53 $\pm$ 0.20          & 57.80 $\pm$ 0.31          \\
            CAL~\cite{zhu2021second}                                         & 91.97 $\pm$ 0.32                                       & 90.93 $\pm$ 0.31                  & 90.75 $\pm$ 0.30          & 90.74 $\pm$ 0.24          & 85.36 $\pm$ 0.16          & 61.73 $\pm$ 0.42 \\
            \hline
            Ours - $p(Y|X)$                                                  & 92.41 $\pm$ 0.25                                       & 91.04 $\pm$ 0.03                  & 91.19 $\pm$ 0.30          & 91.11 $\pm$ 0.45          & 85.67 $\pm$ 0.62          & 59.03 $\pm$ 0.44          \\
            Ours - $p(X|Y)$                                                  & \textbf{92.57 $\pm$ 0.20}                              & \textbf{91.97 $\pm$ 0.09}         & \textbf{91.42 $\pm$ 0.06} & \textbf{91.83 $\pm$ 0.12} & \textbf{86.99 $\pm$ 0.36} & \textbf{62.34 $\pm$ 0.22} \\
            \hline
        \end{tabular}
        \caption{Accuracy (\%) on the test set for CIFAR-10N/100N. Results are taken from~\cite{DBLP:conf/iclr/WeiZ0L0022} using methods containing a single classifier with ResNet-34. Best results are highlighted.
        }
        \label{tab:cifarN}
    \end{table*}

    \begin{table}[t]
        \centering
        \begin{tabular}{ >{\columncolor[HTML]{EFEFEF}}l |cccc}
            \hline
            \cellcolor[HTML]{EFEFEF}                                     & \multicolumn{4}{c}{\cellcolor[HTML]{EFEFEF}Noise rate} \\
            \cline{2-5} \multirow{-2}{*}{\cellcolor[HTML]{EFEFEF}Method} & 0.2                                                   & 0.4            & 0.6            & 0.8            \\
            \hline
            CE                                                           & 47.36                                                 & 42.70          & 37.30          & 29.76          \\
            Mixup~\cite{zhang2017mixup}                                  & 49.10                                                 & 46.40          & 40.58          & 33.58          \\
            Ours - $p(Y|X)$                                              & \textbf{54.53}                                        & \textbf{50.23} & \textbf{45.12} & \textbf{35.78} \\
            Ours - $p(X|Y)$                                              & 53.34                                                 & 49.56          & 44.08          & 34.70          \\
            \hline
            DivideMix~\cite{li2020dividemix}                             & 50.96                                                 & 46.72          & 43.14          & 34.50          \\
            MentorMix~\cite{jiang2020beyond}                             & 51.02                                                 & 47.14          & 43.80          & 33.46          \\
            FaMUS~\cite{xu2021faster}                                    & 51.42                                                 & 48.06          & 45.10          & 35.50          \\
            Ours - $p(Y|X)$ ensemble                                     & \textbf{58.46}                                        & \textbf{53.38} & \textbf{48.92} & \textbf{40.68} \\
            Ours - $p(X|Y)$ ensemble                                     & 57.56                                                 & 52.68          & 47.12          & 39.54          \\
            \hline
        \end{tabular}
        \caption{Test accuracy (\%) on Red Mini-ImageNet with different noise rates and baselines from FaMUS~\cite{xu2021faster}. The upper part shows single-classifier comparisons, and the bottom part shows results for classifier ensembles. Best results are highlighted in both setups.}
        \label{tab:red}
    \end{table}

    \begin{table*}
        [!htb]
        \begin{minipage}{.33\linewidth}
            \label{tab:animal}
            \centering
            \begin{tabular}{ >{\columncolor[HTML]{EFEFEF}}l |c}
                \hline
                Method                                & \cellcolor[HTML]{EFEFEF}Accuracy \\
                \hline
                CE                                    & 79.4                             \\
                SELFIE~\cite{song2019selfie}          & 81.8                             \\
                PLC~\cite{zhang2021learning}          & 83.4                             \\
                Ours - $p(Y|X)$                       & 82.4                             \\
                Ours - $p(X|Y)$                       & \underline{84.7}                 \\
                \hline
                JoCoR~\cite{wei2020combating}         & 82.8                             \\
                Nested + Co-T~\cite{chen2021boosting} & 84.1                             \\
                InstanceGM~\cite{garg2023instance}    & 84.6                             \\
                Ours - $p(Y|X)$ ensemble              & 84.1                             \\
                Ours - $p(X|Y)$ ensemble              & \textbf{85.9}                    \\
                \hline
            \end{tabular}
        \end{minipage}%
        \begin{minipage}{.33\linewidth}
            \centering

            \begin{tabular}{ >{\columncolor[HTML]{EFEFEF}}l |c}
                \hline
                Method                           & \cellcolor[HTML]{EFEFEF}Accuracy \\
                \hline
                CE                               & 68.94                            \\
                Forward~\cite{patrini2017making} & 69.84                            \\
                PTD-R-V~\cite{xia2020part}       & 71.67                            \\
                ELR~\cite{liu2020early}          & 72.87                            \\
                kMEIDTM~\cite{cheng2022instance} & 73.84                            \\
                CausalNL~\cite{yao2021instance}  & 72.24                            \\
                \hline
                Ours - $p(Y|X)$ (ensemble)       & \underline{73.92}                \\
                Ours - $p(X|Y)$ (ensemble)       & \textbf{74.35}                   \\
                \hline
            \end{tabular}
        \end{minipage}
        \begin{minipage}{.33\linewidth}
            \centering

            \begin{tabular}{ >{\columncolor[HTML]{EFEFEF}}l |cc}
                \hline
                Dataset/Methods                                                     & \cellcolor[HTML]{EFEFEF}Webvision & \cellcolor[HTML]{EFEFEF}ImageNet \\
                \hline
                Forward~\cite{patrini2017making}                                    & 61.10                             & 57.36                            \\
                Co-teaching~\cite{han2018co}                                        & 63.58                             & 61.48                            \\
                ELR+~\cite{liu2020early}                                            & \underline{77.78}                 & 70.29                            \\
                UNICON~\cite{karim2022unicon}                                       & 77.60                             & \textbf{75.29}                   \\
                \hline
                \begin{tabular}[c]{@{}l@{}}Ours - $p(Y|X)$\\(ensemble)\end{tabular} & \textbf{78.72}                    & \underline{75.28}                \\
                \begin{tabular}[c]{@{}l@{}}Ours - $p(X|Y)$\\(ensemble)\end{tabular} & 76.32                             & 72.08                            \\
                \hline
            \end{tabular}
        \end{minipage}
        \caption{\textbf{Left}: Test accuracy (\%) on Animal-10N with different noise rates and baselines from InstanceGM~\cite{garg2023instance}. The upper part shows single-classifier comparisons and the bottom part shows results for classifier ensembles. \textbf{Middle}: Test accuracy (\%) on the test set of Clothing1M. Results are obtained from the respective papers; we only use the noisy training set. \textbf{Right}: Top-1 test accuracy (\%) on Mini-WebVision and ImageNet. Results are obtained from the respective papers. Best results are highlighted and second-best results are underlined.}
        \label{tab:animal_c1m_web}
    \end{table*}

    \subsection{Experimental results}
    \label{sec:experimental_results}

    \textbf{Synthetic benchmarks.} The experimental results for instance-dependent noise (IDN) on CIFAR-10/100 are shown in Table~\ref{tab:idn_cifar}.
Compared with the previous state-of-the-art (SOTA) method kMEIDTM~\cite{cheng2022instance}, we achieve competitive performance on CIFAR-10 at low noise rates and improve accuracy by up to 16 percentage points (pp) at high noise rates. On CIFAR-100, we obtain consistent gains of around 3--5 pp across all noise rates. Relative to the prior SOTA generative model, CausalNL~\cite{yao2021instance}, our improvements are substantial at all noise rates. These results indicate that our implicit generative modeling and clean-label prior construction are effective for learning with label noise. Between the two variants $p(X|Y)$ and $p(Y|X)$, ``Ours - $p(X|Y)$'' generally performs better on these benchmarks. Additionally, our approach outperforms existing methods without the need for a dual-network ensembling configuration. This suggests that generative modeling, coupled with iterative EM optimization, can effectively mitigate confirmation bias.

    \textbf{NLP benchmarks.} Experimental results on the NLP datasets are summarized in Tab.~\ref{tab:nlp_news}. Compared with the prior state-of-the-art (SOTA) method DyGEN~\cite{zhuang2023dygen}, our approach attains SOTA performance on both datasets across all noise types and rates. Without NPC post-processing~\cite{bae2022noisy}, we outperform DyGEN by 4--10 pp in every setting. When NPC post-processing is applied, our method achieves an additional $\sim$2 pp improvement over DyGEN. These results indicate that our method learns robustly under label noise across modalities (images and text) and across diverse noise types and rates.

    \textbf{Real-world benchmarks.} In Tab.~\ref{tab:cifarN}, we show the performance of our method on the CIFAR-10N/100N benchmarks. Compared with other single-model baselines, our method achieves an improvement of at least 1 pp across all noise settings on CIFAR-10N and is competitive on CIFAR-100N. Compared with CAL~\cite{zhu2021second}, which uses a complex multi-stage training process, our approach is a much simpler end-to-end alternative. Comparing the two optimization goals, the conclusion for CIFAR-10N/100N matches that for CIFAR-IDN: $p(X|Y)$ generally performs better than $p(Y|X)$.

The Red Mini-ImageNet results in Tab.~\ref{tab:red} show that our method achieves state-of-the-art results for all noise rates, with improvements of about 2 pp with a single model and around 6 pp with a two-model ensemble, compared with prior SOTA methods such as FaMUS~\cite{xu2021faster} and DivideMix~\cite{li2020dividemix}. We observe that $p(Y|X)$ performs best on Red Mini-ImageNet, whereas $p(X|Y)$ performs better on the other datasets. We hypothesize that this is because Red Mini-ImageNet, derived from ImageNet~\cite{deng2009imagenet}, follows a different data-generation process from CIFAR, where images are first web-crawled and then annotated by humans. This implies that the data distribution is more diverse and complex than the label distribution, so $p(Y|X)$ better captures the conditional dependence. Moreover, ImageNet may contain out-of-distribution (OoD) samples that do not belong to any of the predefined classes, which can negatively affect the optimization of $p(X|Y)$, since it tries to fit the data distribution regardless of label quality.

In Tab.~\ref{tab:animal_c1m_web}, we present results on Animal-10N (left), Clothing1M (middle), and Mini-WebVision (right). Our single model on Animal-10N achieves a 1 pp improvement over the single-model baseline, SELFIE~\cite{song2019selfie}. With a two-model ensemble, we improve over the SOTA Nested + Co-T~\cite{chen2021boosting} by 1 pp. For Clothing1M, our ensemble attains a competitive 74.4\%, which is 2 pp better than the previous SOTA generative model, CausalNL~\cite{yao2021instance}. For Mini-WebVision, we outperform the prior semi-supervised SOTA UNICON~\cite{karim2022unicon} by 1 pp in terms of Top-1 accuracy on the WebVision test set and show competitive performance on the ImageNet test set. Regarding the two optimization goals, we observe that $p(X|Y)$ performs better on Animal-10N and Clothing1M, while $p(Y|X)$ performs better on Mini-WebVision. Similar to CIFAR, Animal-10N and Clothing1M are collected based on predefined classes (Clothing1M further discards images if the surrounding captions do not mention any of the 14 classes~\cite{xiao2015learning}). By contrast, Mini-WebVision is known to contain OoD samples due to web collection, which suggests that it is better represented by $p(Y|X)$.

    \begin{table}[]
    \centering
    \setlength{\tabcolsep}{0.8 mm}
    \scalebox{0.86}{
    \begin{tabular}{l|ccccc}
        \hline
        \rowcolor[HTML]{EFEFEF}            & CE   & DivideMix~\cite{li2020dividemix} & CausalNL~\cite{yao2021instance} & InstanceGM~\cite{garg2023instance} & Ours          \\
        \hline
        \cellcolor[HTML]{EFEFEF}CIFAR-100 (50\% IDN) & 2.1h & 7.1h                             & 3.3h                            & 30.5h                              & \textbf{2.3}h \\
        \cellcolor[HTML]{EFEFEF}Clothing1M           & 8h   & 14h                              & 12h                             & 43h                                & \textbf{8.5}h \\
        \hline
    \end{tabular}}
    \caption{Training time (hours) on CIFAR-100 (50\% IDN) and Clothing1M, measured on the hardware described in Sec.~\ref{sec:implementation}.}
    \label{tab:training_time}
\end{table}

    \textbf{Training time comparison.} One of the advantages of our approach is its efficient training, particularly compared with other generative and discriminative methods. Tab.~\ref{tab:training_time} reports training time on CIFAR-100 (50\% IDN) and Clothing1M using the hardware specified in Sec.~\ref{sec:implementation}. Overall, our method achieves shorter training time than competing approaches. It is comparable to training with CE loss and approximately $2\times$ faster than CausalNL~\cite{yao2021instance}, $3\times$ faster than DivideMix~\cite{li2020dividemix}, and $10\times$ faster than InstanceGM~\cite{garg2023instance}.

    \section{Ablation studies}
    \label{sec:ablation}

    \subsection{Transition matrix estimation}
    \label{sec:transition_matrix_estimation}
    \begin{figure*}
        \centering
        \includegraphics[width=0.88\linewidth]{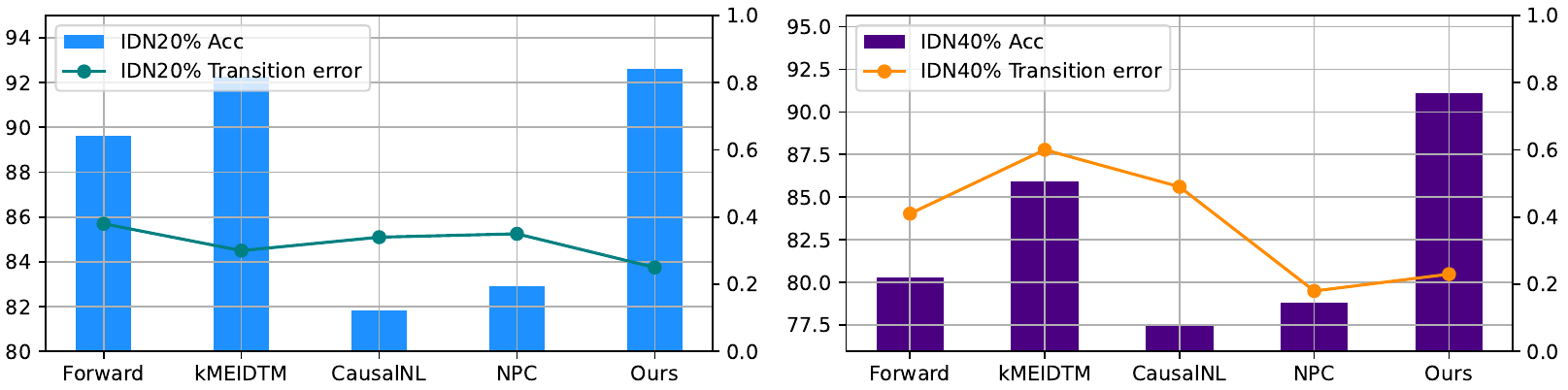}
        \caption{Transition-matrix MSE and classification accuracy on CIFAR-10 IDN at 20\% and 40\% noise rates. Baseline results are taken from NPC~\cite{bae2022noisy} and kMEIDTM~\cite{cheng2022instance}. The right y-axis shows accuracy; the left y-axis shows transition-matrix MSE ($\times$100).}
        \label{fig:transition}
    \end{figure*}

    In Fig.~\ref{fig:transition}, we show the mean squared error (MSE) of transition-matrix estimation and classification accuracy on CIFAR-10 IDN with 20\% and 40\% noise rates, compared with SOTA transition-matrix methods and generative modeling methods. Compared with previous transition-matrix methods, Forward~\cite{patrini2017making} and kMEIDTM~\cite{cheng2022instance}, we observe that our method achieves a lower transition-matrix error and higher accuracy for both noise rates. Compared with the generative methods CausalNL~\cite{yao2021instance} and NPC~\cite{bae2022noisy}, our method achieves comparable transition-matrix estimation error but significantly higher accuracy. This indicates the effectiveness of our generative modeling for instance-dependent transition-matrix estimation.

    \subsection{Hyper-parameter analysis}
\begin{table}[]
    \centering
    \begin{tabular}{ >{\columncolor[HTML]{EFEFEF}}c |cccc}
        \hline
        \cellcolor[HTML]{EFEFEF}                                  & \multicolumn{4}{c}{\cellcolor[HTML]{EFEFEF}CIFAR10-IDN} \\
        \cellcolor[HTML]{EFEFEF}                                  & 20\%  & 30\%  & 40\%  & 50\%   \\
        \hline
        $\beta=0.9$, $K=1$                                        & 92.60 & 91.96 & 91.11 & 90.57  \\
        \hline
        $\beta =0.8$                                              & 92.49 & 91.88 & 90.83 & 88.81 \\
        $\beta =0.7$                                              & 91.55 & 90.87 & 90.62 & 88.40 \\
        $\beta =0.5$                                              & 89.13 & 87.98 & 87.48 & 85.73 \\
        $K=3$                                                     & 92.30 & 91.83 & 90.83 & 89.75 \\
        \hline
    \end{tabular}
    \caption{Test accuracy (\%) on CIFAR-10-IDN to study the hyper-parameters $\beta$ in Eq.~\eqref{eq:moving_average} and $K$ in Eq.~\eqref{eq:ce}.}
    \label{tab:hyper}
\end{table}
\begin{figure}[htbp]
     \centering
     \begin{subfigure}[b]{0.49\columnwidth}
         \centering
         \includegraphics[width=\linewidth]{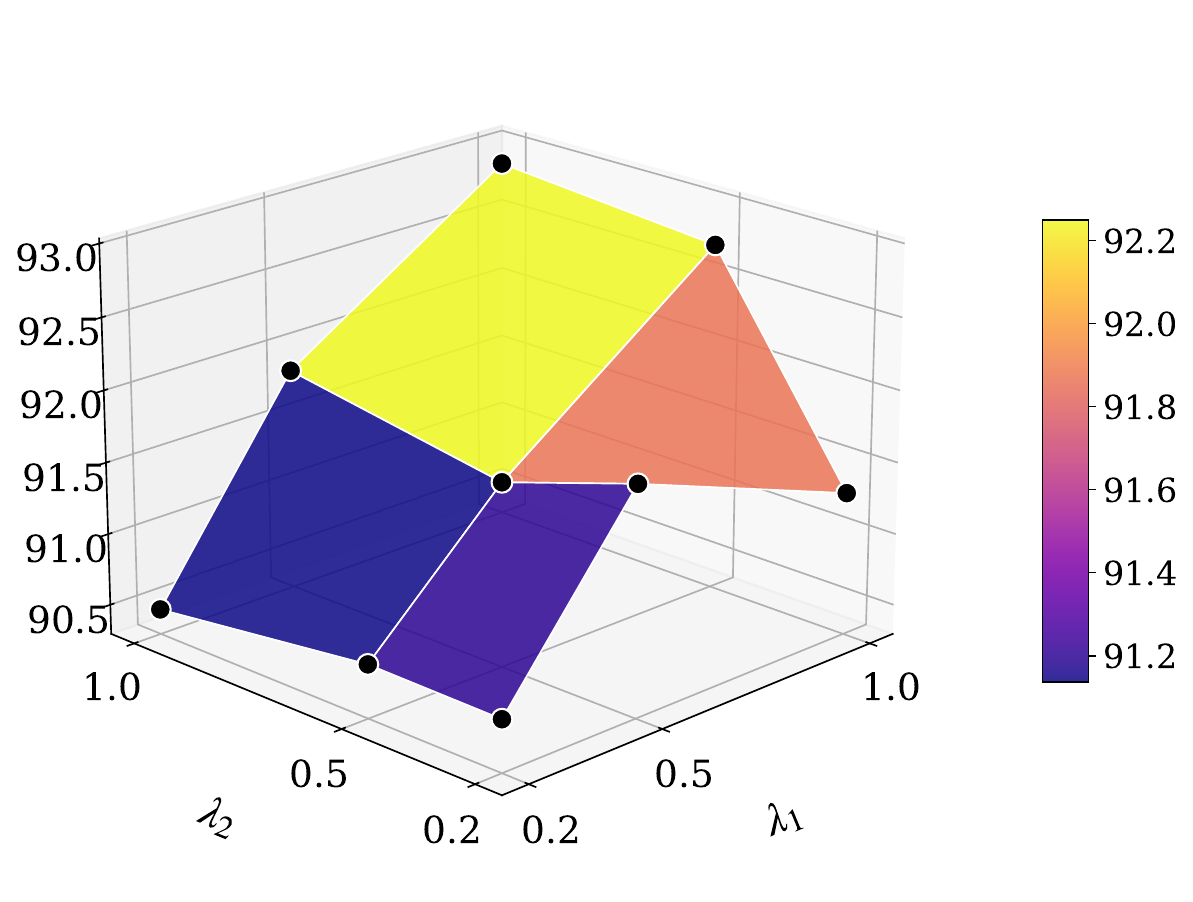}
         \caption{Noise rate $r=0.2$}
         \label{fig:accuracy_r02}
     \end{subfigure}
     \hfill %
     \begin{subfigure}[b]{0.49\columnwidth}
         \centering
         \includegraphics[width=\linewidth]{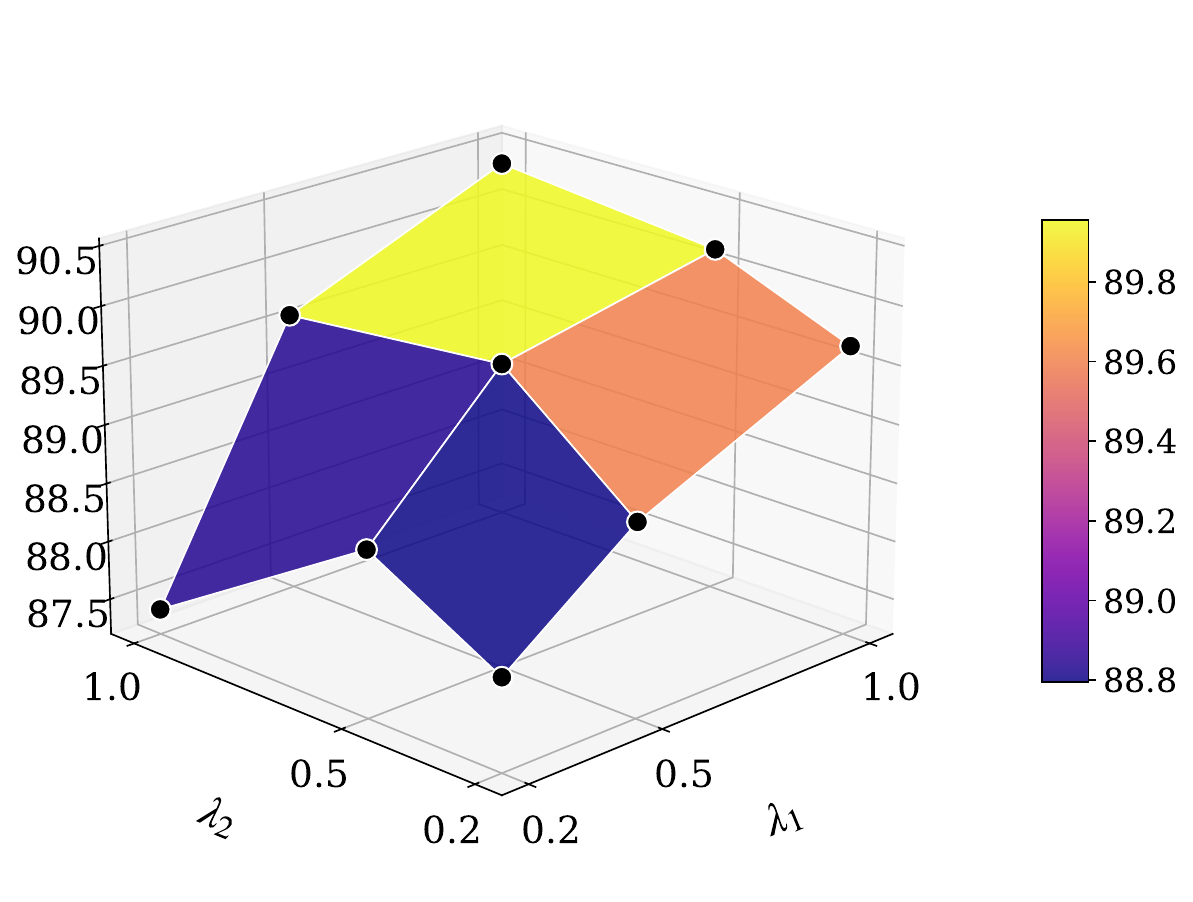}
         \caption{Noise rate $r=0.5$}
         \label{fig:accuracy_r05}
     \end{subfigure}
     
     \caption{Effect of loss weights $(\lambda_1$ and $\lambda_2)$ of $\mathcal{L}_{PRI}$ and $\mathcal{L}_{KL}$ on test accuracy for CIFAR-10-IDN under different noise rates: (a) $r=0.2$ and (b) $r=0.5$. }
     \label{fig:weight_search_comparison}
\end{figure}

    \begin{figure}[htbp]
     \centering
     \begin{subfigure}[b]{0.49\columnwidth}
         \centering
         \includegraphics[width=\linewidth]{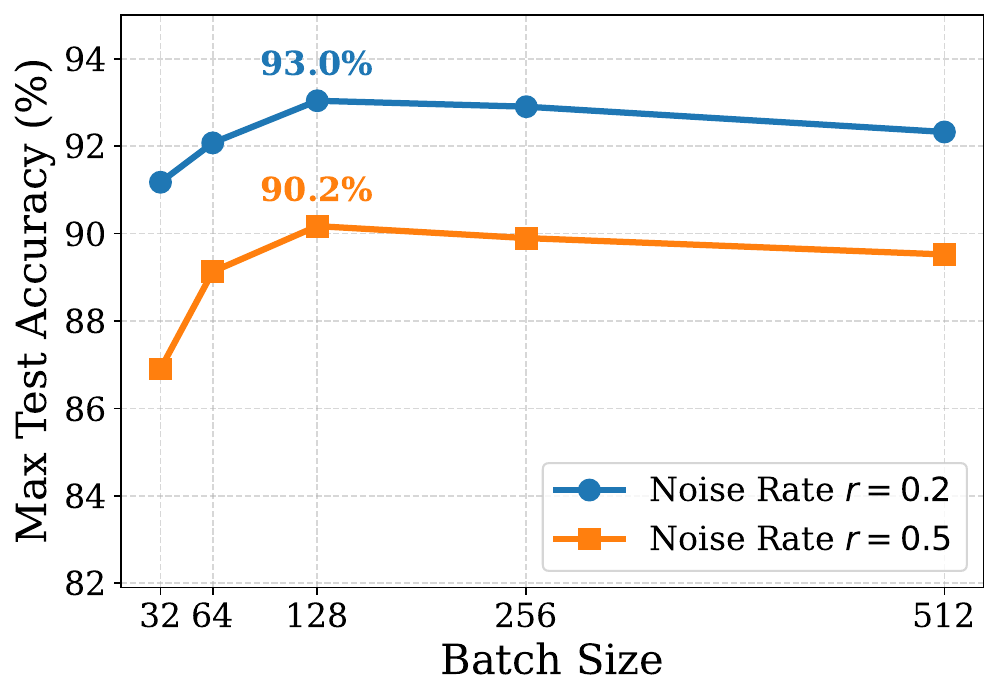}
         \caption{CIFAR-10 IDN}
         \label{fig:accuracy_r02}
     \end{subfigure}
     \hfill %
     \begin{subfigure}[b]{0.49\columnwidth}
         \centering
         \includegraphics[width=\linewidth]{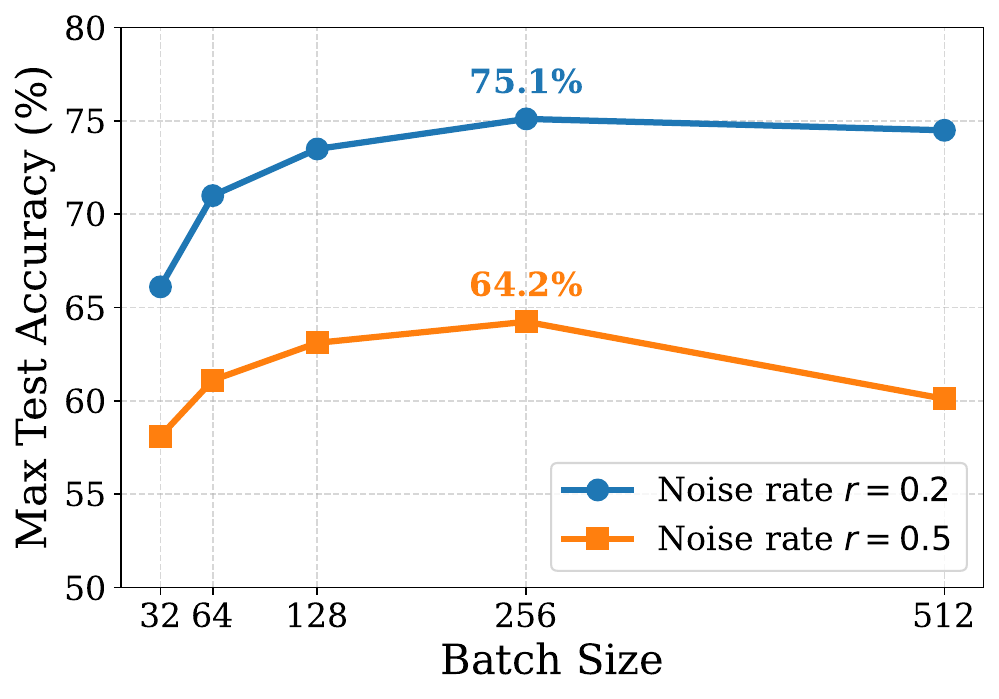}
         \caption{CIFAR-100 IDN}
         \label{fig:accuracy_r05}
     \end{subfigure}
     
     \caption{Effect of batch size on test accuracy for (a) CIFAR-10 and (b) CIFAR-100 IDN under different noise rates: $r=0.2$ and $r=0.5$.}
     \label{fig:batch}
\end{figure}

\begin{table*}[t]
\centering
\small
\setlength{\tabcolsep}{5pt}
\begin{tabular}{lcc|cccc|c}
\toprule
& \multicolumn{2}{c|}{Components} & \multicolumn{4}{c|}{CIFAR-10-IDN Accuracy (\%)} & Avg \\
Setting & Cov & Unc & 20\% & 30\% & 40\% & 50\% & (\%) \\
\midrule
\multicolumn{8}{l}{\textbf{Both components (reference)}} \\
\textbf{coverage + uncertainty (full)} & $\checkmark$ & $\checkmark$
& \textbf{92.60} & \textbf{91.96} & \textbf{91.11} & \textbf{90.57} & \textbf{91.56} \\
\midrule
\multicolumn{8}{l}{\textbf{Coverage-related variants}} \\
coverage only & $\checkmark$ & --
& 84.96 {\scriptsize(-7.64)} & 83.19 {\scriptsize(-8.77)} & 81.88 {\scriptsize(-9.23)} & 78.38 {\scriptsize(-12.19)} & 82.10 {\scriptsize(-9.46)} \\
no EMA ($\beta{=}0$) & $\checkmark$ & $\checkmark$
& 84.57 {\scriptsize(-8.03)} & 81.59 {\scriptsize(-10.37)} & 68.88 {\scriptsize(-22.23)} & 61.47 {\scriptsize(-29.10)} & 74.13 {\scriptsize(-17.43)} \\
no sampling ($\arg\max$) & $\checkmark$ & $\checkmark$
& 20.19 {\scriptsize(-72.41)} & 18.56 {\scriptsize(-73.40)} & 16.09 {\scriptsize(-75.02)} & 15.26 {\scriptsize(-75.31)} & 17.53 {\scriptsize(-74.03)} \\
\midrule
\multicolumn{8}{l}{\textbf{Uncertainty-related variants}} \\
uncertainty only & -- & $\checkmark$
& 85.57 {\scriptsize(-7.03)} & 81.00 {\scriptsize(-10.96)} & 72.42 {\scriptsize(-18.69)} & 66.61 {\scriptsize(-23.96)} & 76.40 {\scriptsize(-15.16)} \\
w/ random $w_i$ & $\checkmark$ & $\checkmark$
& 90.10 {\scriptsize(-2.50)} & 89.66 {\scriptsize(-2.30)} & 86.25 {\scriptsize(-4.86)} & 84.28 {\scriptsize(-6.29)} & 87.57 {\scriptsize(-3.99)} \\
\midrule
\multicolumn{8}{l}{\textbf{Baseline}} \\
noisy labels only (CE) & -- & -- 
& 86.93 {\scriptsize(-5.67)} & 82.42 {\scriptsize(-9.54)} & 76.68 {\scriptsize(-14.43)} & 58.93 {\scriptsize(-31.64)} & 76.24 {\scriptsize(-15.32)} \\
\bottomrule
\end{tabular}
\caption{Ablation about coverage and uncertainty. The number in parentheses shows the difference with respect to our proposed model at each noise rate. The Avg column shows the average results from the other columns, and bold denotes the best result per column.}
\label{tab:abl_grouped}
\end{table*}

    \begin{table}[t]
        \centering
        \scalebox{0.9}{
        \begin{tabular}{ >{\columncolor[HTML]{EFEFEF}}c |ccc|cccc}
            \hline
                        & \cellcolor[HTML]{EFEFEF}                                                                  & \cellcolor[HTML]{EFEFEF}                                               & \cellcolor[HTML]{EFEFEF}                              & \multicolumn{4}{c}{\cellcolor[HTML]{EFEFEF}CIFAR10-IDN} \\
            \cline{5-8} & \multirow{-2}{*}{\cellcolor[HTML]{EFEFEF}$p(\tilde{\mathbf{y}}| \mathbf{y}, \mathbf{x})$} & \multirow{-2}{*}{\cellcolor[HTML]{EFEFEF}$q( \mathbf{y}| \mathbf{x})$} & \multirow{-2}{*}{\cellcolor[HTML]{EFEFEF}E-step $KL$} & \cellcolor[HTML]{EFEFEF}20\%                           & \cellcolor[HTML]{EFEFEF}30\% & \cellcolor[HTML]{EFEFEF}40\% & \cellcolor[HTML]{EFEFEF}50\% \\
            \hline
            \#1         & \checkmark                                                                                &                                                                        &                                                       & 86.93                                                  & 82.43                        & 76.68                        & 58.93                        \\
            \#2         &                                                                                           & Eq.~\eqref{eq:reverse_partial_loss}                                    &                                                       & 43.19                                                  & N/A                          & N/A                          & N/A                          \\
            \#3         & \checkmark                                                                                & Eq.~\eqref{eq:ce_partial_loss}                                         & \checkmark                                            & 85.96                                                  & 82.74                        & 78.34                        & 73.72                        \\
            \#4         & \checkmark                                                                                & Eq.~\eqref{eq:partial_loss}                                            & \checkmark                                            & 91.36                                                  & 90.88                        & 90.23                        & 88.47                        \\
            \#5         & \checkmark                                                                                & Eq.~\eqref{eq:reverse_partial_loss}                                    &                                                       & 92.40                                                  & 90.23                        & 87.85                        & 80.46                        \\
            \#6         & \checkmark                                                                                & Eq.~\eqref{eq:reverse_partial_loss}                                    & \checkmark                                            & 92.60                                                  & 91.96                        & 91.11                        & 90.57                        \\
            \hline
        \end{tabular}}
        \caption{Ablation study of EM formulation. N/A indicates model collapse -- please see Sec.~\ref{sec:loss_fn_analysis} for more details.}
        \label{tab:optima_goal_analysis}
    \end{table}
    \textbf{Hyperparameter sensitivity:} In Tab.~\ref{tab:hyper}, we perform a hyperparameter sensitivity test for our method on CIFAR-10-IDN, including coverage and uncertainty for prior-label construction. To test label coverage, we first examine performance as a function of $\beta \in \{0.5,0.7,0.8,0.9\}$ in Eq.~\eqref{eq:moving_average}, where the default is $\beta=0.9$. We observe that performance does not change much for $\beta \in \{0.7,0.8,0.9\}$, which indicates our model’s robustness with respect to this hyperparameter. For $\beta=0.5$, performance drops significantly, indicating that using a moving average with a relatively high $\beta$ is important to avoid overfitting. We test $K \in \{1,3\}$ in Eq.~\eqref{eq:ce} by sampling $\{\hat{\mathbf{y}}_{i,j}\}_{j=1}^{K} \sim \mathsf{Cat}(g_{\theta}(\mathbf{x}_{i}))$. We observe no significant change in performance with the higher value of $K$; therefore, we choose $K=1$ for simplicity.

    \textbf{Loss weighting:} We also sweep the $\lambda_1$ and $\lambda_2$ over $\{0.2, 0.5, 1.0\}$ on CIFAR-10-IDN under noise rates $r=0.2$ and $r=0.5$. As shown in Fig.~\ref{fig:weight_search_comparison}, the test performance remains relatively stable across different weight combinations, with the equally balanced setting achieving slightly higher accuracy than the others. These results indicate that our model is robust to the choice of individual loss weights.

    \textbf{Batch size sensitivity:} We conduct a batch-size ablation study on CIFAR-10/100 IDN under noise rates $r=0.2$ and $r=0.5$, with results shown in Fig.~\ref{fig:batch}. We observe that smaller batch sizes lead to inferior performance, as the mini-batch approximation of $p(\mathbf{x}\mid\mathbf{y})$ becomes biased when fewer samples are used. In contrast, increasing the batch size yields more pronounced performance improvements on CIFAR-100 than on CIFAR-10, reflecting the higher class complexity of CIFAR-100. However, further increasing the batch size beyond a moderate range does not consistently improve performance. We hypothesize that although larger batches provide a more accurate approximation of $p(\mathbf{x}\mid\mathbf{y})$, they may incur optimization difficulties, e.g., convergence to sharp local minima, as reported in prior work~\cite{keskar2016large}. Overall, the batch size of 128 offers a strong trade-off between approximation accuracy and optimization stability, while a batch size of 256 can further improve performance on CIFAR-100.

    \subsection{EM optimization analysis}
    \label{sec:loss_fn_analysis}
    We show an ablation analysis of our EM formulation on CIFAR-10-IDN in Tab.~\ref{tab:optima_goal_analysis}. 
    
    \begin{itemize}
        \item \textbf{row\#1}: training only the noise-transition layer without regularization from the ELBO or the KL divergence. We observed that model performance drops severely at high noise rates.

        \item \textbf{row\#2}: training directly with partial labels generated using a classifier $q(\mathbf{y}| \mathbf{x})$. We observed that without the EM optimization as guidance, the model collapses with significantly worse results at low noise rates and at high noise rates, it fails to converge, suggesting the importance of our EM optimization.

        \item \textbf{row\#3}: maintaining most of the EM framework, but replacing the label prior loss $\mathcal{L}_{PRI}$ in Eq.~\eqref{eq:partial_loss} with a soft cross-entropy that learns directly from partial labels, defined by:
\begin{equation}
    \begin{split}
        \mathcal{L}_{CE\_PRI}(\theta,\mathcal{D})
        = \frac{1}{|\mathcal{D}|} \sum_{(\mathbf{x}_i, \tilde{\mathbf{y}}_i) \in \mathcal{D}}
        \mathbf{p}_{i} \log g_{\theta}(\mathbf{x}_{i}),
    \end{split}
    \label{eq:ce_partial_loss}
\end{equation}
where $\mathbf{p}_i$ is computed from Eq.~\eqref{eq:true_label_prior}.
This experiment shows that such a replacement causes a performance drop at all noise rates, demonstrating the importance of our proposed $\mathcal{L}_{PRI}$ and the approximation of the generative model $p(\mathbf{x} | \mathbf{y})$.

        \item \textbf{row\#4}: comparing the original KL and the reversed KL implementations for $\mathcal{L}_{PRI}$, explained in Eq.~\eqref{eq:partial_loss} and Eq.~\eqref{eq:reverse_partial_loss}, respectively. 
        Although the results obtained using the original KL divergence are competitive, they are inferior to those achieved with the reverse KL (see subsequent rows). This observation aligns with the argument in~\cite{wang2019symmetric} that the original KL divergence typically converges more effectively but is less robust, whereas the reverse KL offers greater robustness to label noise at the cost of slower or less stable convergence.
        Another important point is that when $\mathbf{p}_{i}$, from Eq.~\eqref{eq:true_label_prior}, is approximately one-hot, it can cause numerical instability in the KL divergence in Eq.~\eqref{eq:partial_loss}.

        \item \textbf{row\#5}: omitting the E-step KL divergence (i.e., maximizing only the ELBO in the M-step). The performance remains similar at low noise ratios but drops at high noise rates, indicating the importance of our EM formulation, and the E-step in particular.

        \item \textbf{row\#6}: training with the complete objective function using Eq.~\eqref{eq:optim_goal} and Eq.~\eqref{eq:reverse_partial_loss}. This row produces the best performance across all noise rates.
    \end{itemize}

    \subsection{Coverage and uncertainty analysis}
    
    \textbf{Coverage.} Table~\ref{tab:abl_grouped} groups variants by coverage- vs. uncertainty-related changes.
To probe the \emph{coverage mechanism}, we (i) disable the moving average by setting $\beta{=}0$, (ii) replace stochastic sampling with a deterministic update that relies entirely on the model’s current prediction (the ``$\arg\max$'' variant), and (iii) remove the uncertainty term while keeping coverage (``coverage only''). All three choices degrade performance relative to our proposed model. In particular, %
$\arg\max$ is the worst, most likely due to the strong confirmation bias from a label prior dominated by a single incorrect label,
$\beta{=}0$ similarly hurts performance by overfitting to transient but inaccurate predictions. The \textit{coverage only} result further shows that the coverage vector $\mathbf{c}_i$ (Eq.~\eqref{eq:metrics}) is necessary but insufficient alone; EMA smoothing and stochastic sampling are critical to stabilize the prior, and uncertainty provides complementary regularization.

	extbf{Uncertainty.} We vary the \emph{uncertainty component} by (i) using uncertainty without coverage (``uncertainty only'') and (ii) replacing learned per-sample weights with random values (``random $w_i$''). Both variants underperform our proposed model, especially at higher noise rates, highlighting that uncertainty must work in tandem with coverage, and that $w_i$ should encode the probability of the $i^{th}$ training sample being clean rather than just being a noise weight. For reference, the standard CE baseline (``noisy labels only (CE)'') lags far behind under heavy noise, reinforcing the benefits of combining coverage with uncertainty.

    \subsection{Coverage and uncertainty visualisation}
    \begin{figure*}[h]
        \centering
        \begin{subfigure}
            [t]{\textwidth}
            \centering
            \includegraphics[width=\textwidth]{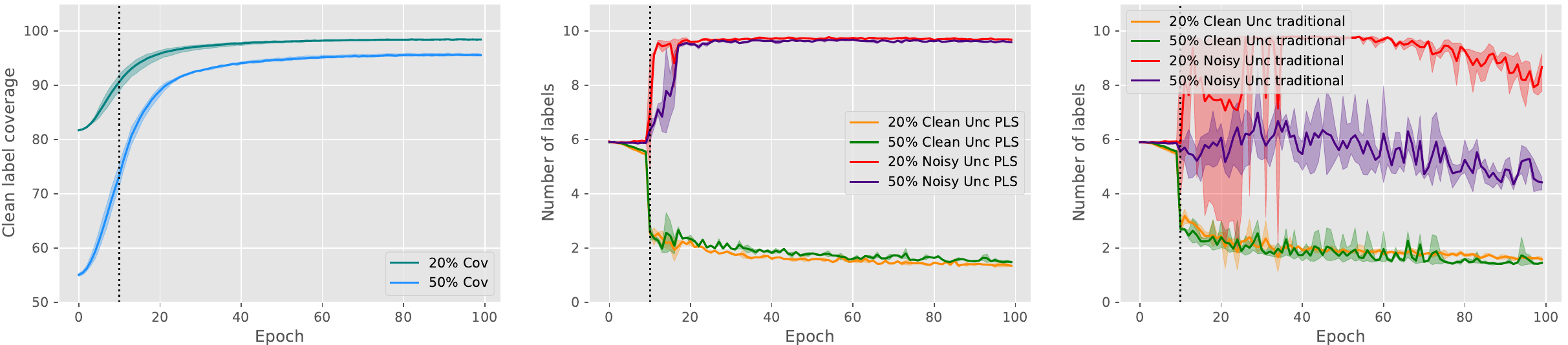}
            \vspace{-15pt}
            \caption{CIFAR10-IDN at 20\% and 50\% noise rates}
            \vspace{10pt}
        \end{subfigure}
        \hfill
        \centering
        \begin{subfigure}
            [t]{\textwidth}
            \centering
            \includegraphics[width=\textwidth]{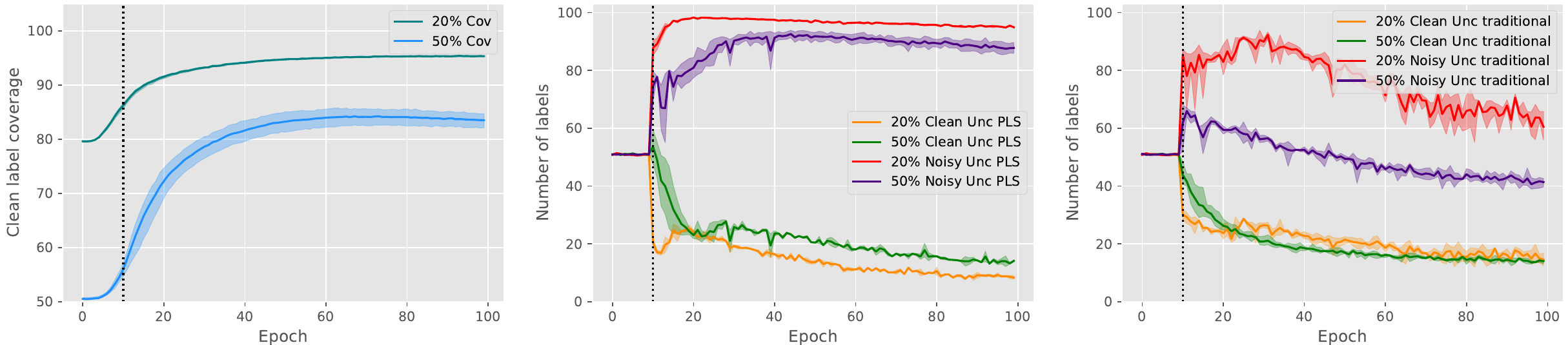}
            \vspace{-15pt}
            \caption{CIFAR100-IDN at 20\% and 50\% noise rates}
            \vspace{10pt}
        \end{subfigure}
        \hfill
        \centering
        \begin{subfigure}
            [t]{\textwidth}
            \centering
            \includegraphics[width=\textwidth]{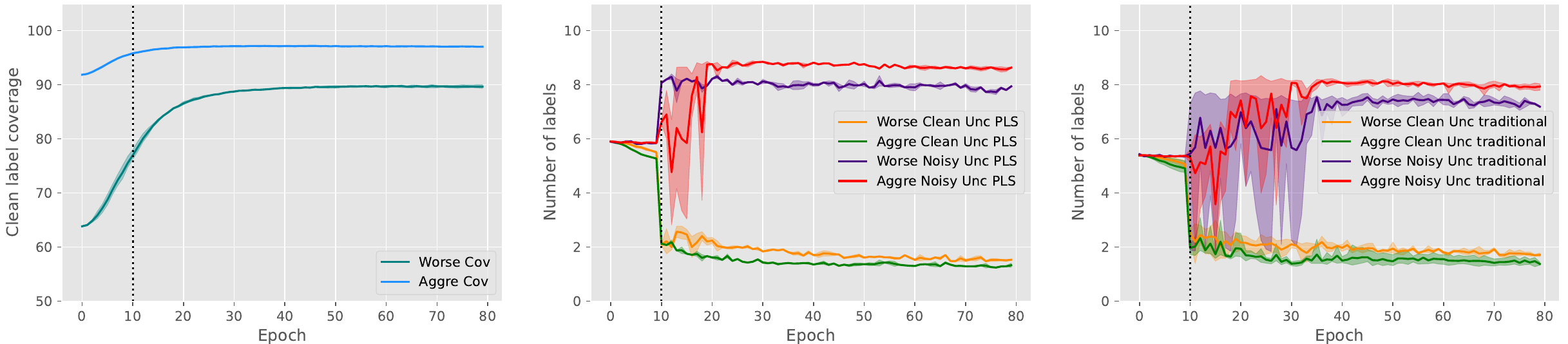}
            \vspace{-15pt}
            \caption{CIFAR10N at "Worse" and "Aggre".}
        \end{subfigure}
        \caption{Visualization of the coverage and uncertainty of our label prior in Eq.~\eqref{eq:metrics}. Coverage (Cov) and Uncertainty (Unc) for \textbf{(a)} CIFAR-10-IDN (20\% and 50\%), \textbf{(b)} CIFAR-100-IDN (20\% and 50\%), and \textbf{(c)} CIFAR-10N (``Worse'' and ``Aggre''). Left-column graphs show clean-label coverage. Middle-column graphs show clean/noisy sample uncertainty under the proposed PLS-trained posterior $q(\mathbf{y}|\mathbf{x})$. The right-column graphs show clean/noisy sample uncertainty when training $q(\mathbf{y}|\mathbf{x})$ without our proposed PLS. The dotted vertical line indicates the end of warm-up training.}
        \label{fig:three graphs}
    \end{figure*}
    We visualize the coverage and uncertainty of our label prior from Eq.~\eqref{eq:metrics} at each training epoch for CIFAR-10/100 with IDN and CIFAR-10N setups in Fig.~\ref{fig:three graphs}. In all cases, label coverage increases as training progresses, indicating that our label prior tends to cover the clean label. In fact, coverage reaches nearly 100\% for CIFAR-10 at 20\% IDN and 96\% for 50\% IDN. For CIFAR-100 at 50\% IDN, we achieve 84\% coverage, and for CIFAR-10N ``Worse,'' we reach 90\% coverage. In terms of clean-label uncertainty and noisy-label uncertainty, we observe a clear gap between different training samples. For clean-label uncertainty, we notice a steady reduction as training progresses for all datasets, which indicates that our $p(\mathbf{y})$ can recover the desired one-hot label for clean-labeled samples. For noisy-label samples, we notice that $p(\mathbf{y})$ has high uncertainty during training, which regularizes the training of noisy-label samples.

In the rightmost column of Fig.~\ref{fig:three graphs}, we plot the clean and noisy-label uncertainty by replacing $\bar{\mathbf{y}}$ (i.e., the class-probability distribution produced by $q(\mathbf{y}|\mathbf{x})$) in Eq.~\eqref{eq:loss_gmm} with the transition output $p(\tilde{\mathbf{y}}| \mathbf{y}, \mathbf{x})$, which follows the traditional small-loss criterion~\cite{li2020dividemix}. This modification affects the calculation of $w_{i}$ in Eq.~\eqref{eq:loss_gmm} and the construction of $p(\mathbf{y})$.

    In this case, we observe highly unstable uncertainty in $p(\mathbf{y})$ for both clean and noisy-label samples. This highlights the importance of our $q(\mathbf{y}|\mathbf{x})$ trained with PLS and its contribution to the construction of sample-wise $p(\mathbf{y})$.
These results suggest that our label-prior distribution effectively selects the correct clean label while reducing the number of label candidates during training.

    \section{Limitations and Solutions}
    \label{sec:discussion}
    
    \textbf{Robustness to a large number of classes.} A core component of our proposed framework is the approximation of the generative distribution $p(\mathbf{x}|\mathbf{y})$ using the variational posterior $q(\mathbf{y}|\mathbf{x})$, as defined in Eq.~\eqref{eq:estimation_p_x_y}. In our current implementation, the denominator $\sum_{i=1}^{|\mathcal{D}|} q(\mathbf{y}| \mathbf{x}_{i})$ is computed within a minibatch of training data instead of the entire training set. While this approach is computationally efficient and has proven effective for the benchmarks used in this paper, it presents a potential limitation when scaling to datasets with a very large number of categories, such as full ImageNet~\cite{deng2009imagenet} or full WebVision~\cite{li2017webvision}, where the number of classes greatly exceeds the size of each batch. For such large-scale datasets, a single minibatch is unlikely to contain representative samples for all classes. This can lead to a sparse and unstable estimation of the data distribution, potentially hindering performance by providing an inaccurate and biased denominator for the $p(\mathbf{x}|\mathbf{y})$ approximation. Furthermore, when scaling up the current single batch strategy to a sliding-window strategy, the model will store multiple outdated predictions that differ dramatically from the current model state, causing inaccurate predictions and making training dynamics unstable.

To overcome the limitations of a sliding-window approach, particularly the issue of storing outdated and inconsistent predictions, we propose integrating a momentum encoder, a technique successfully employed in self-supervised contrastive learning~\cite{he2020momentum}. The denominator in our approximation of $p(\mathbf{x}|\mathbf{y})$ can be computed using this momentum encoder, which maintains a consistent set of representations over time. The momentum encoder's parameters are a slowly evolving moving average of the primary encoder's parameters. This update strategy ensures that the encoded representations remain stable even as the primary model changes rapidly during training. This consistency allows us to leverage a key technique from self-supervised contrastive learning: a large memory bank. By decoupling the number of negative samples from the minibatch size, the memory bank provides a much richer and more stable sampling of the data distribution, leading to a more accurate approximation of $p(\mathbf{x}|\mathbf{y})$.

\textbf{Balanced vs.\ imbalanced dataset distributions.} Our current experimental setups mainly focus on standard noisy-label learning benchmarks, where all datasets have balanced class distributions. If the target dataset is not balanced, such as Clothing1M~\cite{xiao2015learning} or WebVision~\cite{li2017webvision}, a common practice is to re-sample the dataset to make it pseudo-balanced~\cite{li2020dividemix}. However, in real-world scenarios, datasets tend to be imbalanced and entangled with noisy-label annotations. This poses challenges for extending our approach to these settings. Below we list the limitations and potential solutions:

1) \emph{Valid estimation of $w_{i}$ (probability a sample is noisily labeled).} In the standard balanced setup, noise is the primary confounder and our approach works well: $q(\mathbf{y}|\mathbf{x})$ learns to approximate the clean-label distribution, making the loss a good indicator for estimating $w_{i}$. Under long-tailed distributions, however, the training loss is also affected by majority or minority status, reducing the reliability of $w_{i}$. This could lead to higher $w_{i}$ values for clean-labeled samples from minority classes, which also exhibit relatively higher loss. Our approach would need to explore alternative ways to estimate $w_{i}$.
2) \emph{Uniform-distribution assumption for $\mathbf{u}_{i}$.} In the balanced setup, all labels are treated equally with a uniform sampling rate. In long-tailed distributions, the clean-label prior should reflect the data distribution; majority classes should have higher prior probability mass than minority classes. The uniform assumption can be problematic here, but it is relatively straightforward to replace it with a class-weighted prior using known imbalance ratios.
3) \emph{Adjusting $p(\tilde{\mathbf{y}}| \mathbf{y}, \mathbf{x})$.} This term is implemented via a class-dependent noise transition matrix that implicitly assumes uniformity. Under long-tailed distributions, the transition matrix should also incorporate class imbalance. Furthermore, the associated cross-entropy loss is affected by class imbalance; this can be mitigated by incorporating class imbalance ratios into logit adjustment~\cite{menon2020long}.
Overall, several components require adjustment to adapt to the long-tailed noisy-label problem, and we leave a comprehensive treatment to future work.

    \section{Conclusion}

We presented a direction-agnostic EM framework that bridges generative and discriminative noisy-label learning by implicitly estimating the causal ($Y \rightarrow X$) or anti-causal ($X \rightarrow Y$) generation processes  
with a discriminative model, allowing us to eliminate the need for training inefficient generative models.  
Furthermore, we introduce Partial Label Supervision (PLS), which constructs an informative, sample-wise clean-label prior by jointly controlling coverage and uncertainty. Experimental results on vision and NLP benchmarks with diverse noise types/rates show that our method achieves state-of-the-art accuracy, reduces training time, and yields lower transition-matrix error. %
Extensive ablations illustrate that both our coverage and uncertainty approaches are necessary and effective; the method is robust to hyper-parameters, and a simple theoretical analysis explains why high-coverage partial labels reduce worst-case error versus hard re-labeling.

\section*{Acknowledgements}

This work was supported by the Engineering and Physical Sciences Research Council (EPSRC) through grant EP/Y018036/1 and Australian Research Council through grant DP180103232/FT190100525.
The authors would like to thank the editor and the anonymous reviewers for their valuable comments and suggestions.

    \bibliography{egbib}
    \bibliographystyle{abbrv}

    \section{Biography Section}

    \vspace{-33pt}
    \begin{IEEEbiography}[{\includegraphics[width=1in,height=1.25in,clip,keepaspectratio]{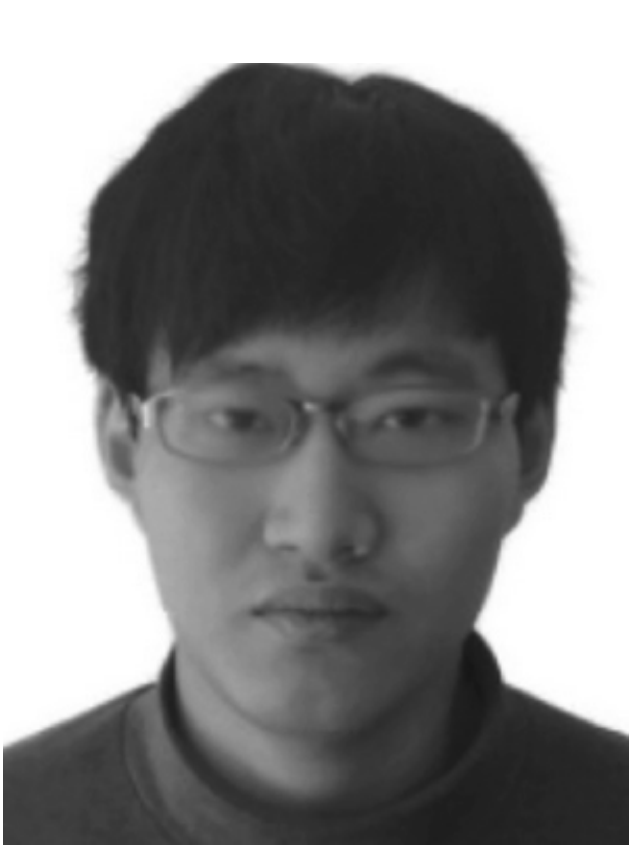}}]{Fengbei Liu} is a Postdoctoral Researcher at Cornell Tech, Cornell University. He received the Ph.D. degree in computer science from the Australian Institute for Machine Learning (AIML), University of Adelaide. His research interests are computer vision and weakly-supervised learning.
    \end{IEEEbiography}
    \vspace{-33pt}
    \begin{IEEEbiography}[{\includegraphics[width=0.93in,height=1.25in,clip,keepaspectratio]{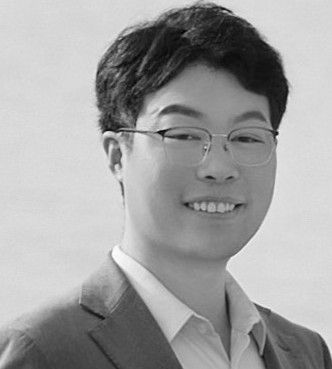}}]{Chong Wang} is a Postdoctoral Researcher at the Department of Radiology, Stanford University. He received the Ph.D. degree in computer science from the Australian Institute for Machine Learning (AIML), University of Adelaide. His research mainly focuses on trustworthy AI, foundation models, machine learning, and computer vision.
    \end{IEEEbiography}
    \vspace{-33pt}
    \begin{IEEEbiography}[{\includegraphics[width=0.96in,height=1.25in,clip,keepaspectratio]{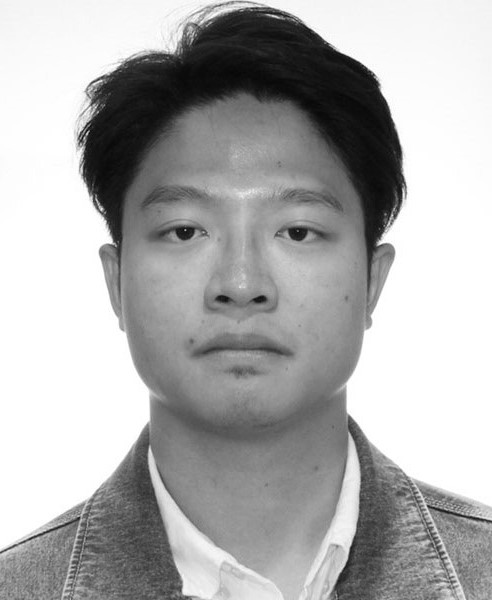}}]{Yuanhong Chen} is a Postdoctoral Researcher at the Australian Institute for Machine Learning (AIML), University of Adelaide. He received the Ph.D. degree in computer science from the Australian Institute for Machine Learning (AIML), University of Adelaide. His research interests are multi-modal learning, computer vision, and medical imaging analysis.
    \end{IEEEbiography}
    \vspace{-33pt}
    \begin{IEEEbiography}[{\includegraphics[width=1.0in,height=1.25in,clip,keepaspectratio]{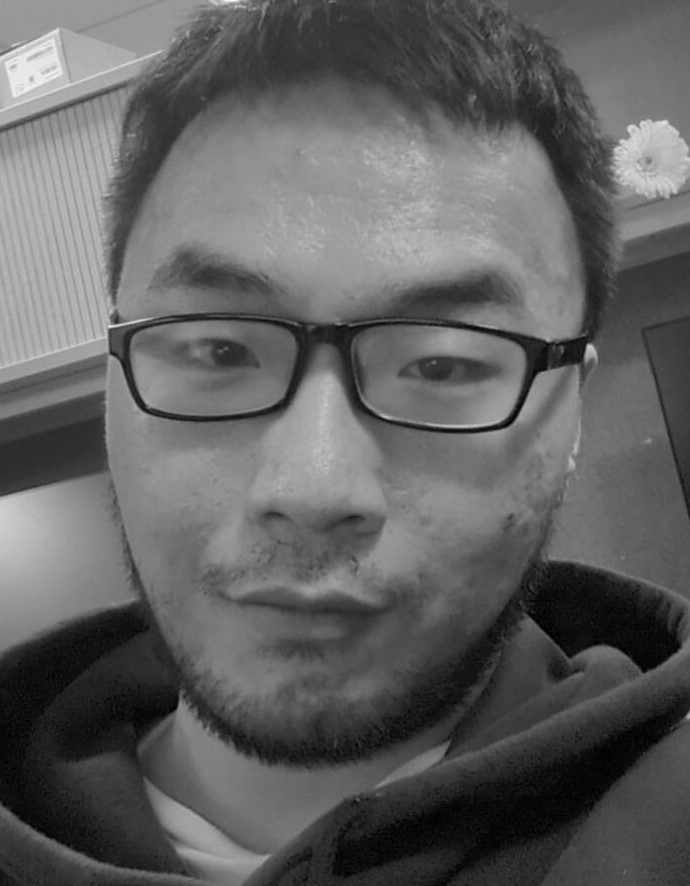}}]{Yuyuan Liu} is a Postdoctoral Researcher at Oxford University. He received the Ph.D. degree in computer science from the Australian Institute for Machine Learning (AIML), University of Adelaide. His research interests are machine learning and computer vision.
    \end{IEEEbiography}
    \vspace{-33pt}
    \begin{IEEEbiography}[{\includegraphics[width=0.97in,height=1.25in,clip,keepaspectratio]{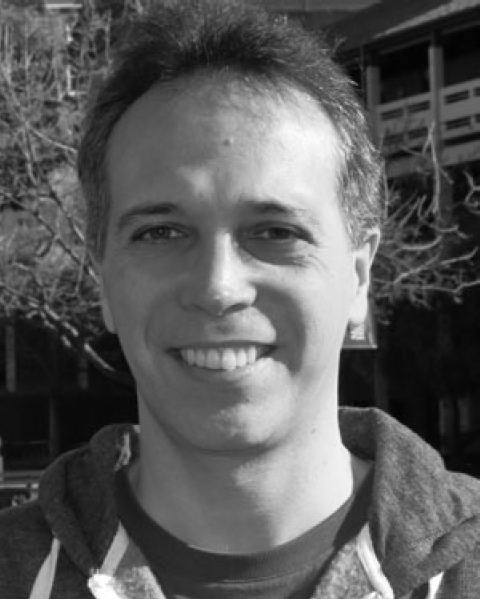}}]{Gustavo Carneiro} is currently a Full Professor of AI and Machine Learning at the Centre for Vision, Speech and Signal Processing (CVSSP), The University of Surrey, UK. 
He received the Ph.D. degree in computer science from the University of Toronto in 2004. 
Previously, he was a professor at the University of Adelaide, an ARC Future Fellow, and the Director of Medical Machine Learning at the Australian Institute of Machine Learning (AIML). 
He joined the University of Adelaide as a senior lecturer in 2011 and became an associate professor in 2015 and a professor in 2019. 
In 2014 and 2019, he joined the Technical University of Munich as a visiting professor and a Humboldt fellow. 
From 2008 to 2011 he was a Marie Curie IIF fellow and a visiting assistant professor with the Instituto Superior Tecnico (Lisbon, Portugal) within the Carnegie Mellon University-Portugal program (CMU-Portugal).
From 2006 to 2008, he was a research scientist with Siemens Corporate Research in Princeton, USA. 
In 2005, he was a postdoctoral fellow with the University of British Columbia and with the University of California San Diego. 
His research interests lie in the fields of computer vision, medical image analysis, and machine learning.
    \end{IEEEbiography}

    \vfill
\end{document}